\documentclass[journal,twocolumn]{IEEEtran}
\usepackage{subfigure}

\usepackage{graphicx}
\usepackage{url}
\usepackage{pslatex}
\usepackage{bm}
\usepackage{amsmath,amssymb,amsthm}  
\usepackage{paralist}

\usepackage{algorithm}
\usepackage{algorithmic}        
\usepackage{multirow}
\renewcommand{\algorithmicrequire}{\textbf{Initialization:}}   
\renewcommand{\algorithmicensure}{\textbf{Iteration:}}
\renewcommand{\algorithmiclastcon}{\textbf{Output:}} 

\newcommand{\e}{\begin{equation}}
\newcommand{\ee}{\end{equation}}
\newcommand{\en}{\begin{equation*}}
\newcommand{\een}{\end{equation*}}
\newcommand{\eqn}{\begin{eqnarray}}
\newcommand{\eeqn}{\end{eqnarray}}
\newcommand{\bmat}{\begin{bmatrix}}
\newcommand{\emat}{\end{bmatrix}}
\newcommand{\BIT}{\begin{itemize}}
\newcommand{\EIT}{\end{itemize}}

\newcommand{\vb}{\bm b}
\newcommand{\vc}{\bm c}

\newcommand{\ve}{\bm e}

\newcommand{\vq}{\bm q}

\newcommand{\vu}{\bm u}
\newcommand{\vv}{\bm v}

\newcommand{\vx}{\bm x}
\newcommand{\vy}{\bm y}

\newcommand{\vgamma}{\bm\gamma}

\newcommand{\mA}{\bm A}
\newcommand{\mB}{\bm B}

\newcommand{\mF}{\bm F}

\newcommand{\mH}{\bm H}
\newcommand{\mI}{\bm I}

\newcommand{\mQ}{\bm Q}
\newcommand{\mR}{\bm R}

\newcommand{\mU}{\bm U}
\newcommand{\mV}{\bm V}

\newcommand{\mSigma}{\bm \Sigma}

\newcounter{oursection}

\ifCLASSINFOpdf
\else
\fi

\usepackage{subfigure}
\usepackage{hyperref}
\usepackage{epstopdf}
\AppendGraphicsExtensions{.tif}
\hypersetup{
	colorlinks=true,
	linkcolor=blue, 
	filecolor=magenta,      
	urlcolor=cyan,
}

\begin{document}
	%
	\title{Acceleration of RED via  Vector Extrapolation}
	
	\author{Tao Hong, Yaniv Romano and Michael Elad
		
		\thanks{T. Hong and M. Elad are with the Department
			of Computer Science, Technion - Israel Institute of Technology, Haifa, 32000, Israel (e-mail:\{hongtao,elad\} @cs.technion.ac.il).}
		\thanks{Y. Romano is with the Department of Statistics, Stanford University, Stanford, CA 94305, U.S.A. (email: yromano@stanford.edu).}
	}	
	\maketitle
	
	\begin{abstract}
		Models play an important role in inverse problems, serving as the prior for  representing the original signal to be recovered. REgularization by Denoising
		(RED) is a recently introduced general framework for constructing such priors using state-of-the-art denoising algorithms. Using RED, solving inverse problems is shown to amount to an iterated denoising process. However, as the complexity of denoising algorithms is generally high, this might lead to an overall slow algorithm. 
		In this paper, we suggest an accelerated technique based on vector extrapolation (VE) to speed-up existing RED solvers. Numerical experiments validate the obtained gain by VE, leading to {\color{black}a substantial  savings in computations compared with the original fixed-point method}. 
	\end{abstract}

	%
	\IEEEpeerreviewmaketitle

  \section{Introduction}\label{Sec:Intro}
Inverse problems in imaging address the reconstruction of clean images from their corrupted versions. The corruption can be a blur, loss of samples, downscale or a more complicated operator (e.g., CT and MRI), accompanied by a noise contamination. Roughly speaking, inverse problems are characterized by two main parts: the first is called the forward model, which formulates the relation between the noisy measurement and the desired signal, and the second is the prior, describing the log-probability of the destination signal. 
	
In recent years, we have witnessed a massive advancement  in a basic inverse problem referred to as image denoising  \cite{buades2005non,elad2006image,dabov2007image,   zoran2011learning,dong2013nonlocally,chen2017trainable}. Indeed, recent work goes as far as speculating that the performance obtained by leading image denoising algorithms is getting very close to the possible ceiling \cite{chatterjee2010denoising,milanfar2013tour,levin2011natural}. This motivated researchers to seek ways to exploit this progress in order to address general inverse problems. Successful attempts, as in \cite{protter2009generalizing,danielyan2012bm3d,metzler2015optimal}, suggested an exhaustive manual adaptation of existing denoising algorithms, or the priors used in them, treating  specific alternative missions. This line of work has a clear limitation, as it does not offer a flexible and general scheme for incorporating various image denoising achievements for tackling other advanced image processing tasks. 
This led to the following natural question: is it possible to suggest a general framework that utilizes the abundance of high-performance image denoising algorithms for addressing general inverse problems? Venkatakrishnan et al. gave a positive answer to this question, proposing a framework called Plug-and-Play Priors ($P^3$) method \cite{venkatakrishnan2013plug,sreehari2016plug,chan2017plug}. Formulating the inverse problem as an optimization task and handling it via the Alternating Direction Method of Multipliers (ADMM) scheme \cite{boyd2011distributed}, $P^3$ shows that the whole problem is decomposed into a sequence of image denoising sub-problems, coupled with simpler computational steps. The $P^3$ scheme provides a constructive answer to the desire to use denoisers within inverse problems, but it suffers from several key disadvantages: $P^3$ does not define a clear objective function, since the regularization used is implicit; Tuning the parameters in $P^3$ is extremely delicate; and since $P^3$ is tightly coupled with the ADMM, it has no flexibility with respect to the numerical scheme. 

A novel framework named REgularization by Denoising (RED) \cite{romano2017little} proposes an appealing alternative while overcoming all these flaws. The core idea in RED is the use of the given denoiser within an expression of regularization that generalizes a Laplacian smoothness term. The work in \cite{romano2017little} carefully shows that the gradient of this regularization is in fact the denoising residual. This, in turn, leads to several iterative algorithms, all guaranteed to converge to the global minimum of the inverse problem's penalty function, while using a denoising step in each iteration. 

The idea of using a state-of-the-art denoising algorithm for constructing an advanced prior for general inverse problems is very appealing.\footnote{\color{black}Firstly, it enables utilizing the vast progress in image denoising for solving challenging inverse problems as explained above. Secondly,  RED enables utilizing the denoiser as a black-box.} However, a fundamental problem still exists due to  the high complexity of typical denoising algorithms, which are required to be activated many times in such a recovery process. Indeed, the evidence from the numerical experiments posed in \cite{romano2017little} clearly exposes this problem, in all the three methods proposed, namely the steepest descent, the fixed-point (FP) strategy and the ADMM scheme. Note that the FP method is a parameter free and the most efficient among the three, and yet this approach too requires the activation of the denoising algorithms dozens of times for a completion of the recovery algorithm.


{\color{black} Our main contribution in this paper is to address these difficulties by applying vector extrapolation (VE) \cite{cabay1976,rajeevan1992vector,sidi2017vector} to accelerate the FP algorithm shown in \cite{romano2017little}. Our simulations illustrate the effectiveness of VE for this acceleration, saving more than $50\%$ of the overall computations involved compared with the native FP method.}
	
	
The rest of this paper is organized as follows. We review RED and its  FP method in Section \ref{Sec:Prelim}. Section \ref{Sec:PropMethod} recalls the Vector Extrapolation acceleration idea. Several experiments on image deblurring and super-resolution, which follows the ones given in \cite{romano2017little}, show the effectiveness of VE, and these are brought in Section \ref{Sec:Results}. We conclude our paper in Section \ref{Sec:Conclusion}.


\section{REgularization by Denoising (RED)}\label{Sec:Prelim}
This section reviews the framework of RED, which utilizes denoising algorithms as image priors \cite{romano2017little}. We also describe its original solver based on the Fixed Point (FP) method.  
\subsection{Inverse Problems as Optimization Tasks}\label{subsec:pre:optitasks}
From an estimation point of view, the signal $\vx$ is to be recovered from its measurements $\vy$ using the posterior conditional probability $P(\vx|\vy)$. Using maximum a posterior probability (MAP) and the Bayes' rule, the estimation task is formulated as:
$$
\begin{array}{rcl}
{\vx}^*_{MAP}& =& \arg\max_{\vx}P(\vx|\vy)\\
&=&\arg\max_{\vx} \frac{P(\vy|\vx)P(\vx)}{P(\vy)}\\
&=&\arg\max_{\vx}P(\vy|\vx)P(\vx)\\
&=&\arg\min_{\vx}-\log\{P(\vy|\vx)\}-\log{P(\vx)}.
\end{array}
$$
The third equation is obtained by the fact that $P(\vy)$ does not depend on $\vx$. The term $-\log\{P(\vy|\vx)\}$ is known as the log-likelihood $\ell(\vy,\vx)$. A typical example is
\e
\ell(\vy,\vx)\triangleq -\log\{P(\vy|\vx)\}=\frac{1}{2\sigma^2}\|\mH\vx-\vy\|_2^2\label{eq:datafidelty}
\ee
{\color{black}referring to the case $\vy = \mH \vx +\ve $, where $\mH$ is any linear degradation operator and $\ve$ is a white mean zero Gaussian noise with variance $\sigma^2$.} Note that the expression $\ell(\vy,\vx)$ depends on the distribution of the noise.\footnote{White Gaussian noise is assumed throughout this paper.} Now, we can write the MAP optimization problem as
\e
\vx^*_{MAP}=\arg\min_{\vx}\ell(\vy,\vx)+\alpha R(\vx)\label{eq:MAP:opt}
\ee
where $\alpha>0$ is a trade-off parameter to balance $\ell(\vy,\vx)$ and $R(\vx)$. 
$R(\vx)\triangleq - \log{P(\vx)}$ refers to the prior that describes the statistical nature of $\vx$. This term is typically referred to as the regularization, as it is used to stabilize the inversion by emphasizing the features of the recovered signal. In the following, we will describe how RED activates denoising algorithms for composing $R(\vx)$. Note that Equation \eqref{eq:MAP:opt} defines a wide family of inverse problems including,  but not limited to, inpainting, deblurring, super-resolution, \cite{kirsch2011introduction} and more.

\subsection{RED and the Fixed-Point Method}\label{subsec:pre:RED:model:property:FP}
Define $f(\vx)$ as an abstract and differentiable denoiser.\footnote{This denoising function admits a noisy image $\vx$, and removes additive Gaussian noise form it, assuming a prespecified noise energy.} RED suggests applying the following form as the prior: 
\e
R(\vx) =\frac{1}{2}\vx^\mathcal T\left(\vx-f(\vx)\right),\label{eq:prior:red}
\ee
where $\mathcal T$ denotes the transpose operator. {\color{black} The term $ \vx^\mathcal T\left(\vx - f(\vx)\right)$ is an image-adaptive Laplacian regularizer, which favors either a small residual $\vx - f(\vx)$, or a small inner product between $\vx$ and the residual \cite{romano2017little}.} Plugged into Equation \eqref{eq:MAP:opt}, this leads the following minimization task: 
\e
\min_{\vx} E(\vx)\triangleq \ell(\vy,\vx)+\alpha\frac{1}{2}\vx^\mathcal T\left(\vx-f(\vx)\right).\label{eq:opt:red:abs}
\ee
The prior $R(\vx)$ of RED is a convex function and easily differentiated if the following two conditions are met:
\begin{itemize}
\item Local Homogeneity:  For any scalar $c$ arbitrarily close to $1$, we have $f(c\vx)=cf(\vx)$.
\item Strong Passivity: The Jacobian $\nabla_{\vx}f(\vx)$ is stable in the sense that its  spectral radius is upper bounded by one, $\rho(\nabla_{\vx}f(\vx))\leq 1$.
\end{itemize}
Surprisingly, the gradient of $E(\vx)$ is given by
\e
\nabla_{\vx} E(\vx)= \nabla_{\vx}\ell(\vy,\vx)+\alpha\left(\vx-f(\vx)\right).\label{eq:opt:red:abs:gradient}
\ee
As discussed experimentally and theoretically in \cite[Section 3.2]{romano2017little}, many of  the state-of-the-art denoising algorithms satisfy the above-mentioned two conditions, and thus the gradient of \eqref{eq:opt:red:abs} is simply evaluated through \eqref{eq:opt:red:abs:gradient}. As a consequence, $E(\vx)$ in Equation \eqref{eq:opt:red:abs} is a convex function if $\ell(\vx,\vy)$ is convex, such as in the case of \eqref{eq:datafidelty}. In such cases any gradient-based algorithm can be utilized to address Equation \eqref{eq:opt:red:abs} leading to its global minimum.


Note that evaluating the gradient of $E(\vx)$ calls for one denoising activation, resulting in an expensive operation as the complexity of good-performing denoising algorithms is typically high. Because of the slow convergence speed of the steepest descent and the high complexity of ADMM, the work reported in \cite{romano2017little} suggested using the FP method to handle the minimization task posed in Equation \eqref{eq:opt:red:abs}. The development of the FP method is rather simple, relying on the fact that the global minimum of \eqref{eq:opt:red:abs} should satisfy the first-order optimality condition, i.e., $\nabla_{\vx}\ell(\vy,\vx)+\alpha\left(\vx-f(\vx)\right)=\bm 0$. For the FP method, we utilize the following iterative formula to solve this equation:
\e
\nabla_{\vx}\ell(\vy,\vx_{k+1})+\alpha \left(\vx_{k+1}-f(\vx_k)\right)=\bm 0.\label{eq:FP:recursive:first-order}
\ee
The explicit expression of the above for $\ell(\vy,\vx)=\frac{1}{2\sigma^2}\|\mH\vx-\vy\|_2^2$ is
\e
\vx_{k+1}=\left[\frac{1}{\sigma^2}\mH^\mathcal T\mH +\alpha \mI \right]^{-1}\left[\frac{1}{\sigma^2}\mH ^\mathcal T\vy +\alpha f(\vx_k)\right],\label{eq:FP:recursive:explicit}
\ee
where $\mI$ represents the identity matrix.\footnote{\color{black} This matrix inversion is calculated in the Fourier domain for block-circulant $\mH$, or using iterative methods for the more general cases.} The convergence of the FP method is guaranteed since
$$
\rho\left( \left[\frac{1}{\sigma^2}\mH^\mathcal T\mH +\alpha \mI \right]^{-1} \alpha \nabla_{\vx} f(\vx_k)\right)< 1.
$$

Although the FP method is more efficient than the steepest descent and the ADMM, it still needs hundreds of iterations, which means hundreds of denoising activations, to reach the desired minimum. This results in a high complexity algorithm which we aim to address in this work. In the next section, we introduce an accelerated technique called Vector Extrapolation (VE) to substantially reduce the amount of iterations in the FP method.
\section{Proposed Method via Vector Extrapolation}\label{Sec:PropMethod}
We begin this section by introducing the philosophy of VE in linear and nonlinear systems and then discuss three variants of VE\footnote{{\color{black} We refer the interesting readers to \cite{sidi2017vector} and the references therein to explore further the VE technique.}}, i.e., Minimal Polynomial Extrapolation (MPE), Reduced Rank Extrapolation (RRE) and Singular Value Decomposition Minimal Polynomial Extrapolation (SVD-MPE) \cite{sidi2015svd,sidi2017vector}. Efficient implementation of these three variants is also discussed. We end this section by embedding VE in the FP method for RED, offering an acceleration of this scheme. Finally, we discuss the convergence and stability properties of VE. 
\subsection{VE in Linear and Nonlinear Systems}
Consider a vector set $\{ \vx_i\in\Re^{N}\}$ generated via a linear process,
\e
\vx_{i+1} = \mA\vx_i+\vb,~~ i=0,1,\cdots,\label{eq:FP:linear:proce} 
\ee
where $\mA \in\Re^{N \times N}$, $\vb \in\Re^N$ and $\vx_0$ is the initial vector. If $\rho(\mA)<1$, a limit point $\vx^*$ exists, being the FP of \eqref{eq:FP:linear:proce}, $\vx^*=\mA\vx^*+\vb$. We turn to describe how VE works on such linear systems \cite{smith1987extrapolation}. Denote 
$ \vu_i=\vx_{i+1}-\vx_i,~~i=0,1,\cdots,$ and define the defective vector $\ve_i$ as 
\e
\ve_i = \vx_i-\vx^*,~~ i=0,1,\cdots. \label{eq:defec:vec}
\ee
Subtracting $\vx^*$ from both sides of \eqref{eq:FP:linear:proce} and utilizing the fact that $\vx^*$ is the FP, we have $\ve_{i+1} = \mA \ve_i$ resulting in 
\e
\ve_{i+1} = \mA^{i+1} \ve_0.\label{eq:error:propa:0_i}
\ee
We define a new extrapolated vector $\vx_{(m,\kappa)}$ as a weighted average of the form 
\e
\vx_{(m,\kappa)}=\sum\limits_{i=0}^{\kappa}\gamma_i\vx_{m+i},\label{eq:approxi:weight}
\ee 
where $\sum\limits_{i=0}^{\kappa}\gamma_i = 1$.
Substituting \eqref{eq:defec:vec} in \eqref{eq:approxi:weight} and using \eqref{eq:error:propa:0_i} and $\sum_{i=0}^{\kappa}\gamma_i=1$, we have
\e
\begin{array}{rcl}
	\vx_{(m,\kappa)}&=&\sum\limits_{i=0}^{\kappa} \gamma_i\left(\vx^*+\ve_{m+i}\right)\\
	&=&\vx^*+\sum\limits_{i=0}^{\kappa}\gamma_i\ve_{m+i}\\
	&=&\vx^*+\sum\limits_{i=0}^{\kappa}\gamma_i\mA^{i}\ve_m .\label{eq:approx:recursive}
\end{array}
\ee
Note that the optimal $\{\gamma_i\}$ and $\kappa$ should be chosen so as to force $\sum\limits_{i=0}^{\kappa}\gamma_i\mA^i\ve_{m}=0$. This way, we attain the FP through only one extrapolation step. 
	
More broadly speaking, given a nonzero matrix $\mB\in\Re^{N\times N}$ and an arbitrary nonzero vector  $\vu\in\Re^N$, we can find a unique polynomial $P(z)$ with smallest degree to yield $P(\mB)\vu=\bm 0$. Such a $P(z)$ is called the minimal polynomial of $\mB$ with respect to the vector $\vu$. Notice that the zeros of $P(z)$ are the eigenvalues of $\mB$. Thus, assume that the minimal polynomial of $\mA$ with respect to $\ve_m$ can be represented as 
\e
P(z)=\sum_{i=0}^{\kappa}c_i z^i,~~c_{\kappa}=1 \label{eq:minipolynomialAssum}
\ee
resulting in $P(\mA)\ve_m=\bm 0$. So, we have 	
\e
\sum\limits_{i=0}^{\kappa} c_i{\mA}^i\ve_m = \sum\limits_{i=0}^{\kappa} c_i\ve_{m+i}=\bm 0. \label{eq:polyno:A} 
\ee
Multiplying  both sides of \eqref{eq:polyno:A} by $\mA$ results in $\sum_{i=0}^{\kappa}c_i\mA \ve_{m+i}=\sum_{i=0}^{\kappa}c_i\ve_{m+i+1}=\bm 0$, and thus we receive 
\e
\sum\limits_{i=0}^{\kappa} c_i\ve_{m+i}=\sum_{i=0}^{\kappa}c_i\ve_{m+i+1}=\bm 0.\label{eq:errorEquaLinear}
\ee
Subtracting these expressions gives
\e
\begin{array}{rcl}
\sum\limits_{i=0}^{\kappa} c_i\left(\ve_{m+i+1}-\ve_{m+i}\right) &=&\sum\limits_{i=0}^{\kappa} c_i\left(\vx_{m+i+1}-\vx_{m+i}\right) \\
		&=&\sum\limits_{i=0}^{\kappa} c_i \vu_{m+i} = \bm 0.	
\end{array}
\label{eq:linear:system:c_i}
\ee
This suggests that $\{ c_i \}$ could be determined by solving the linear equations posed in \eqref{eq:linear:system:c_i}. Once obtaining $\{c_i\}$, $\{ \gamma_i \}$ are calculated through $\gamma_i=\frac{c_i}{\sum_{j=0}^{\kappa}c_j}$. Note that $\sum_{j=0}^{\kappa} c_j\neq 0$ if $\mI-\mA$ is not singular yielding $\sum_{j=0}^{\kappa} c_j =P(1)\neq 0$. Assuming $\kappa$ is the degree of the minimal polynomial of $\mA$ with respect to $\ve_m$, we can find a set of $\{ \gamma_i \}$ to satisfy $\sum_{i=0}^{\kappa}\gamma_i=1$ resulting in $\sum_{i=0}^{\kappa}\gamma_i \vx_{m+i}=\vx^*$. However, the degree of the minimal polynomial of $\mA$ can be as large as $N$, which in our case is very high. Moreover, we also do not have a way to obtain this degree with an easy algorithm. Because of these two difficulties, some approximate methods are developed to extrapolate the next vector via the previous ones and we will discuss them in Subsection \ref{subsec:van_MPE}. 

Turning to the nonlinear case, denote $\bm F$ as the FP function to evaluate the next vector,
\begin{equation}
\vx_{i+1} = \bm F(\vx_i),~i=0,1,\cdots, \label{Sec:PropMethod:fixed_define}
\end{equation} 
where $\bm F$ is an $N$-dimensional vector-valued function, $\bm F: \Re^{N}\rightarrow \Re^N$. We say $\vx^*$ is a FP of $\mF$ if $\vx^*=\bm F(\vx^*)$. Expanding $\bm F(\vx)$ in its Taylor series yields
\begin{equation*} 
\bm F(\vx)=\bm F(\vx^*)+\bm F'(\vx^*)(\vx-\vx^*)+O(\|\vx-\vx^*\|^2)~~\text{as}~~\vx\rightarrow \vx^*,
\end{equation*}
where $\bm F'(\cdot)$ is the Jacobian matrix of $\bm F(\cdot)$. Recalling $\bm F(\vx^*)=\vx^*$, we have 
\begin{equation*}
	\bm F(\vx)=\vx^*+\bm F'(\vx^*)(\vx-\vx^*)+O(\|\vx-\vx^*\|^2)~~\text{as}~~\vx\rightarrow \vx^*.
\end{equation*}
Assuming the sequence $\vx_0,\vx_1,\dots$ converges to $\vx^*$ (if $\rho(\mF'(\vx))<1$), it follows that $\vx_i$ will be close enough to $\vx^*$ for all large $i$, and hence
\begin{equation*}
	\vx_{i+1}=\vx^*+\bm F'(\vx^*)(\vx_i-\vx^*)+O(\|\vx_{i}-\vx^*\|^2),~~\text{as}~~i\rightarrow \infty.
\end{equation*}
Then, we rewrite this in the form
\begin{equation*}
	\vx_{i+1}-\vx^*=\bm F'(\vx^*)(\vx_i-\vx^*)+O(\|\vx_{i}-\vx^*\|^2),~~\text{as}~~i\rightarrow \infty.
\end{equation*}
For large $i$, the vectors $\{\vx_i\}$ behave as in the linear system of the form $(\bm I-\bm A)\vx=\bm b$ through 
\begin{equation*}
\vx_{i+1}=\bm A\vx_i+\bm b, ~~ i=0,1,\cdots,
\end{equation*}
where $\bm A=\bm F'(\vx^*)$, $\bm b=[\bm I-\bm F'(\vx^*)]\vx^*$. This implies that the nonlinear system yields the same formula as the linear one and motivates us to extrapolate the next vector by the previous ones as in linear systems. Indeed, such an extension has been shown to be successful in various areas of science and engineering, e.g., computational fluid dynamics, semiconductor research, tomography and geometrical image processing \cite{rajeevan1992vector,rosman2009efficient,sidi2017vector}.

\subsection{Derivations of MPE, RRE and SVD-MPE}\label{subsec:van_MPE}
We turn to discuss how to utilize an approximate way to obtain the next vector by extrapolating the previous ones. Due to the fact that the degree of the minimal polynomial can be as large as $N$ and we cannot obtain it, an arbitrary positive number is set as the degree, being much smaller than the true one. With such a replacement, the linear equations in \eqref{eq:linear:system:c_i} become inconsistent and there does not exist a solution for  $\{ c_i \}, c_{\kappa}=1$ in the ordinary sense. Alternatively, we solve instead
\e
\min_{\vc} \|\mU_{\kappa}^m \vc\|_2^2,~~\text{s.t.} ~~c_{\kappa}=1\label{eq:MPE:LS}
\ee
where $\vc=\bmat c_0&\cdots &c_{\kappa}\emat^\mathcal T$ and $\mU_{\kappa}^m=\bmat \vu_{m}&\cdots&\vu_{m+\kappa}\emat$. Then evaluating $\gamma_i$ through ${c_i}/\left(\sum_{i=0}^{i=\kappa}c_i\right)$ results in the next vector $\vx_{(m,\kappa)}=\sum_{i=0}^{\kappa}\gamma_i \vx_{m+i}$ as a new approximation. This method is known as Minimal Polynomial Extrapolation (MPE) \cite{sidi1991}. 

The detailed steps for obtaining the next vector through MPE are shown in {\bf Algorithm \ref{alg:MPE_Van}}. To solve the constrained problem in \eqref{eq:MPE:LS}, we suggest utilizing QR decomposition with the modified Gram-Schmidt (MGS) \cite{sidi1991,golub2012matrix}. The MGS procedure for the matrix $\bm U_{\kappa}^m$ is shown in {\bf Algorithm \ref{alg:MGS_imple}}. 

\begin{algorithm}[!htb]        
\caption{Minimal Polynomial Extrapolation (MPE)}           
\label{alg:MPE_Van}                 
\begin{algorithmic}[1]
\REQUIRE ~\\
A sequence of vectors $\{\vx_m,\vx_{m+1},\vx_{m+2},\cdots,\vx_{m+\kappa+1}\}$ is produced by the baseline algorithm (FP in our case).	
\lastcon ~\\          
A new vector $\vx_{(m,\kappa)}$.\\
\STATE Construct the matrix
			$$\bm U_{\kappa}^m=\bmat \vx_{m+1}-\vx_m ,\cdots,\vx_{m+\kappa+1}-\vx_{m+\kappa}\emat \in \Re^{N\times (\kappa+1)}$$ and then compute its QR factorization via {\bf Algorithm \ref{alg:MGS_imple}}, $\bm U_{\kappa}^m=\bm Q_{\kappa}\bm R_{\kappa}$.\label{alg:MPE_Van:QR}
\STATE {Denote $\bm r_{{\kappa}+1}$ as the $\kappa+1$th column of $\bm R_{\kappa}$ without the last row and solve the following ${\kappa}\times {\kappa}$ upper triangular system
\begin{equation*}
\bm R_{{\kappa}-1}\bm c'=-r_{{\kappa}+1}~~~ \bm c'=\bmat c_0,c_1,\cdots,c_{{\kappa}-1}\emat^\mathcal T
\end{equation*} where $\bm R_{{\kappa}-1}$ is the previous ${\kappa}$ columns of $\bm R_{\kappa}$ without the last row. Finally, evaluate $\{\gamma_i\}$ through $\{\frac{c_i}{\sum_{i=0}^kc_i}\}$.} \label{alg:MPE_Van:backward}
\STATE {Compute $\bm \xi=\bmat \xi_0,\xi_1,\cdots,\xi_{{\kappa}-1}\emat^\mathcal T$ through 
\begin{equation*}
\xi_0=1-\gamma_0;~~\xi_j=\xi_{j-1}-\gamma_j,~~j=1,\cdots,{\kappa}-1
\end{equation*}}
\STATE Compute $\bm \eta =\bmat\eta_0,\eta_1,\cdots,\eta_{{\kappa}-1} \emat^\mathcal T=\bm R_{{\kappa}-1}\bm \xi$. Then we attain $\vx_{(m,\kappa)} =\vx_m+\bm Q_{{\kappa}-1}\bm \eta$  as the new initial vector where $\bm Q_{{\kappa}-1}$ represents the previous $\kappa$ columns of $\bm Q_{\kappa}$.
\end{algorithmic}
\end{algorithm}
	
\begin{algorithm}[!htb]        
	\caption{Modified Gram-Schmidt (MGS)}           
	\label{alg:MGS_imple}                 
	\begin{algorithmic}[1]
  	\lastcon $\mQ_\kappa$ and $\mR_\kappa$ ($r_{ij}$ denotes the $(i,j)$th element of $\mR_\kappa$ and $\vq_i$ and $\vu_i$ represent the $i$th column of $\mQ_\kappa$ and $\mU_\kappa^m$, respectively.).
  \STATE{Compute $r_{11}=\|\bm u_1\|_2$ and $\bm q_1=\bm u_1/r_{11}$. }
  \FOR{$i=2,\cdots,{\kappa}+1$}
  \STATE Set $\bm u_i^{(1)}=\bm u_{i}$
  \FOR{$j=1,\cdots,i-1$}
  \STATE $r_{ji}=\bm q_j^\mathcal T\bm u_i^{(j)}$ and $\bm u_i^{(j+1)}=\bm u_i^{(j)}-r_{ji}\bm q_j$
 \ENDFOR
\STATE {Compute $r_{ii}=\|\bm u_{i}^{(i)}\|_2$ and $\bm q_i=\bm u_i^{(i)}/r_{ii}$}
\ENDFOR
\end{algorithmic}
\end{algorithm}
	
Now, let us discuss the other two variants of VE, i.e., Reduced Rank Extrapolation (RRE) \cite{sidi1991} and SVD-MPE \cite{sidi2015svd}. The main difference among RRE, MPE and SVD-MPE is at Step \ref{alg:MPE_Van:backward} in {\bf Algorithm \ref{alg:MPE_Van}} regarding the evaluation of $\{\gamma_i\}$. In RRE and SVD-MPE, we utilize the following methods to obtain $\{ \gamma_i \}$:
\begin{itemize}
\item[RRE:]
Solving $\bm R_{\kappa}^\mathcal T\bm R_{\kappa} \bm d = \bm 1$ through forward and backward substitution, we obtain $\vgamma$ through $\frac{\bm d}{\sum_i d_i}$. Actually, such a formulation of $\vgamma$ is the solution of:
\e
\min_{\vgamma} \|\mU_{\kappa}^m \vgamma\|_2^2,~~\text{s.t.} ~~\sum_i\gamma_i=1.\label{eq:RRE:LS}
\ee
\item[SVD-MPE:]
Computing the SVD decomposition of  $\bm R_{\kappa}=\mU\mSigma\mV^\mathcal T$, we have $\vgamma=\frac{\vv_{\kappa+1}}{\sum_i \vv_{i,\kappa+1}}$ where $\vv_{\kappa+1}$ and $\vv_{i,\kappa+1}$ represent the last column and the $(i,\kappa+1)$th element of matrix $\mV$, respectively.
\end{itemize}
Here are two remarks regarding RRE, MPE and SVD-MPE:
\begin{itemize}
\item Observing the derivations of MPE, SVD-MPE and RRE, we notice that RRE's solution must exist unconditionally, while MPE and SVD-MPE may not exist because the sum of $\{c_i \}$ and $\{ v_{i,\kappa+1} \}$ in MPE and SVD-MPE may become zero. Thus RRE may be more robust in practice \cite{rosman2009efficient}. However, MPE and RRE are related, as revealed in \cite{sidi2017minimal}. Specifically, if MPE does not exist, we have $\vx_{(m,\kappa)}^{RRE}=\vx_{(m,\kappa-1)}^{RRE}$. Otherwise, the following holds
\en
\mu_{\kappa} \vx_{(m,\kappa)}^{RRE}=\mu_{\kappa-1}\vx_{(m,\kappa-1)}^{RRE}+v_\kappa\vx_{(m,\kappa)}^{MPE},~~\mu_\kappa=\mu_{\kappa-1}+v_\kappa
\een
where $\mu_\kappa$, $\mu_{\kappa-1}$ and $v_\kappa$ are positive scalars depending only on $\vx_{(m,\kappa)}^{RRE}$, $\vx_{(m,\kappa-1)}^{RRE}$ and $\vx_{(m,\kappa)}^{MPE}$, respectively. Furthermore, the performance of MPE and RRE is similar -- both of the methods either perform well or work poorly \cite{sidi2017vector}. 
\item Observe that we only need to store ${\kappa}+2$ vectors in memory at all steps in {\bf Algorithm \ref{alg:MGS_imple}}. Formulating the matrix $\bm U_m^{\kappa}$, we overwrite the vector $\vx_{m+i}$ with $\bm u_{m+i}=\vx_{m+i}-\vx_{m+i-1}$ when the latter is computed and only $\vx_m$ is always in the memory. Next, $\bm u_{m+i}$ is overwritten by $\bm q_i$, $i=1,\cdots,{\kappa}+1$ in computing the matrix $\bm Q_{\kappa}$. Thus, we do not need to save the vectors $\vx_{m+1},\cdots,\vx_{m+{\kappa}+1}$, which implies that no additional memory is required in running {\bf Algorithm \ref{alg:MGS_imple}}.
\end{itemize}
\subsection{Embedding VE in the Baseline Algorithm}\label{SubSec:Embed:VE} 
We introduce VE in its cycling formulation for practical usage. One cycling means we activate the baseline algorithm to produce $\{\vx_i\}$ and then utilize VE once to evaluate the new vector as a novel initial point. Naturally, we repeat such a cycling many times. The steps of utilizing VE in its cycling mode are shown in {\bf Algorithm \ref{alg:VE:Prac}}. Few comments are in order:

\begin{itemize}
\item In practice, we utilize VE in its cycling formulation. Specially, the iterative form shown in {\bf Algorithm \ref{alg:VE:Prac}} is named as full cycling \cite{sidi2017vector}. To save the complexity of computing $\{\gamma_i\}$, one may reuse the previous $\{\gamma_i\}$, a method known as cycling with frozen $\gamma_i$. Parallel VE can be also developed if more machines are available. Explaining details of the last two strategies is out of the scope of this paper. We refer the reader to \cite{sidi2017vector} for more information. 
\item Numerical experience also indicates that cycling with even moderately large $m>0$ will avoid stalling from happening \cite{sidi1998upper}. Moreover, we also recommend setting $m>0$ when the problem becomes challenging to solve. 
\item In our case, the termination criterion in {\bf Algorithm \ref{alg:VE:Prac}} can be the number of total iterations (the number of calling of the baseline algorithm) or the difference between consecutive two vectors. Furthermore, we also recommend giving additional iterations to activate the baseline algorithm after terminating the VE, which can stabilize the accelerated algorithm in practice. 
\end{itemize}

\begin{algorithm}[!htb]        
	\caption{Baseline Algorithm $+$ Vector Extrapolation}           
		\label{alg:VE:Prac}                 
		\begin{algorithmic}[1]
			\REQUIRE ~\\
			Choose nonnegative integers $m$ and $\kappa$ and an initial vector $\vx_0$. The baseline algorithm is the FP method, as given in \eqref{eq:FP:recursive:explicit}. 
			\lastcon ~\\          
			Final Solution $\vx^*$.\\
			\WHILE {1}
			\STATE Obtain the series of $\vx_i$ through the baseline algorithm where $1\leq i\leq m+\kappa+1$, and save $\vx_{m+i}$ for $0\leq i\leq \kappa+1$ to formulate $\mU_m^\kappa$. \label{alg:VE:Prac:series}\\
			\STATE Call {\bf Algorithm \ref{alg:MPE_Van}} to obtain $\vx_{(m,\kappa)}$.
			\STATE If the termination of the algorithm is satisfied, set $\vx^*=\vx_{(m,\kappa)}$ and break, otherwise, set $\vx_{(m,\kappa)}$ as the new initial point $\vx_0$ and go to Step \ref{alg:VE:Prac:series}. 
			\ENDWHILE
		\end{algorithmic}
	\end{algorithm}

\subsection{Convergence and Stability Properties}\label{SubSec:Conv:Stab}
We mention existing results regarding the convergence and stability properties of VE for understanding this technique better. A rich literature has examined the convergence and stability properties of RRE, MPE and SVD-MPE in linear systems \cite{sidi1986acceleration,sidi1988convergence}. Assuming the matrix $\mA$ is diagonalizable, then in the $k$th iteration $\vx_k$ should have the form $\vx_k=\vx^*+\sum_{i=1}^\kappa \vv_i\lambda_i^k$ where $(\lambda_i,\vv_i)$ are some or all of the eigenvalues and corresponding eigenvectors of $\mA$, with distinct nonzero eigenvalues. By ordering $\lambda_i$ as $|\lambda_1|\geq|\lambda_2|\geq\cdots$, the following asymptotic performance holds for all of the three variants of VE when $|\lambda_k|>|\lambda_{k+1}|$:
\e
\vx_{(m,\kappa)}-\vx^*=O(\lambda_{\kappa+1}^m)~~ \textbf{as} ~~ m\rightarrow \infty.\label{eq:linear:asym:relationship}
\ee
This implies that the sequence $\{\vx_{(m,\kappa)}\}_{m=0}^{\infty}$ converges to $\vx^*$ faster than the original sequence $\{\vx_k\}$.

As shown in \eqref{eq:linear:asym:relationship}, for a large $m$, \eqref{eq:FP:linear:proce} reduces the contributions of the smaller $\lambda_i$ to the error $\vx_{(m,\kappa)}-\vx^*$, while VE eliminates the contributions of the $\kappa$ largest $\lambda_i$. This indicates that $\vx_{(m,\kappa)}-\vx^*$ is smaller than each of the errors $\vx_{m+i}-\vx^*,~i=0,1,\cdots,\kappa$, when $m$ is large enough. We mention another observation that an increasing $\kappa$ generally results in a faster convergence of VE. However, a large $\kappa$ has to increase the storage requirements and also requires a much higher computational cost. Numerical experiments indicate that a moderate $\kappa$ can already works well in practice. 

If the following condition is held, we say VE is stable:
\e
\sup_{m} \sum_{i=0}^\kappa |\gamma_i^{(m,\kappa)}|<\infty.\label{eq:VE:stab:eq}
\ee
Here, we denote $\{\gamma_i\}$ by $\{\gamma_i^{(m,\kappa)}\}$ to show their dependence on $m$ and $\kappa$. If \eqref{eq:VE:stab:eq} holds true, the error in $\vx_i$ will not magnify severely. As shown in \cite{sidi1986acceleration,sidi1988convergence,sidi1986convergence}, MPE and RRE obey such a stability property.

For nonlinear systems, the analysis of convergence and stability becomes extremely challenging. One of the main results is the quadratic convergence theorem \cite{skelboe1980computation,smith1987extrapolation,jbilou1991some}. This theorem is built on one special assumption that $\kappa$ is set to be the degree of the minimal polynomial of $\mF'(\vx^*)$. The proof of the quadratic convergence was shown in \cite{skelboe1980computation}. In a following work, Smith, Ford and Sidi noticed that there exists a gap in the previous proof \cite{smith1987extrapolation}. Jbilou et al. suggested two more conditions in order to close the gap \cite{jbilou1991some}:
\begin{itemize}
\item The matrix $\mF'(\vx^*)-\mI$ is nonsingular 
\item $\mF'(\cdot)$ satisfies the following Lipschitz condition:
$$
\|\mF'(\vx)-\mF'(\vy)\|\leq L \|\vx-\vy\|~~L>0.
$$
\end{itemize}
Surprisingly, these two conditions are met by the RED scheme. The first condition is satisfied by the fact
$$\rho\left( \left[\frac{1}{\sigma^2}\mH^\mathcal T\mH +\alpha \mI \right]^{-1} \alpha \nabla_{\vx} f(\vx)\right)< 1.$$ The second one is also true, due to the assumption in RED that the denoiser $f(\vx)$ is differentiable. So we claim that it is possible for VE to solve RED with quadratic convergence rate.

Although VE can lead to a quadratic convergence rate, trying to achieve such a rate may not be realistic because $\kappa$ can be as large as $N$. However, we may obtain a linear but fast convergence in practice with even moderate values of $m$ and $\kappa$, which is also demonstrated in the following numerical experiments.

\section{Experimental Results}\label{Sec:Results}
We follow the same experiments of image deblurring and super-resolution as presented in \cite{romano2017little} to investigate the performance of VE in acceleration. The trainable nonlinear reaction diffusion (TNRD) method \cite{chen2017trainable} is chosen as the denoising engine. Mainly, we choose the FP method as our baseline algorithm. For a fair comparison, the same parameters suggested in \cite{romano2017little} for different image processing tasks are set in our experiments. {\color{black} In \cite{romano2017little}, the authors compared RED with other popular algorithms in image deblurring and super-resolution tasks, showing its superiority. As the main purpose in this paper is to present the acceleration of our method for solving RED, we omit the comparisons with other popular algorithms.} In the following, we mainly show the acceleration of applying MPE with FP for solving RED first and then discuss the choice of parameters in VE, i.e., $m$ and $\kappa$.  {\color{black} In addition, we compare our method with three other methods, steepest descent (SD), Nesterov's acceleration \cite{nesterov2018lectures} and Limited-memory BFGS (L-BFGS) \cite{jorge2006numerical}. Note that we need to determine a proper step-size for the above methods \cite{jorge2006numerical}. However, evaluating the objective value or gradient in RED is expensive implying that any line-search method becomes prohibitive. Note that, in contrast, in our framework as described in {\bf Algorithm \ref{alg:VE:Prac}} does not suffer from such a problem. In the following, we manually choose a fixed step-size for getting a good convergence behavior.} Finally, we compare the difference among RRE, MPE and SVD-MPE. All of the experiments are conducted on a workstation with Intel(R) Xeon(R) CPU E5-2699 @2.20GHz.
%
\subsection{Image deblurring}\label{SubSec:ImageDeblur}
\begin{figure*}[!htb]
    \centering
   \subfigure[SD: $22.56\text{dB (input)}\rightarrow 25.65\text{dB} \rightarrow  26.47\text{dB} \rightarrow 26.97\text{dB}\rightarrow 27.35\text{dB} \rightarrow \text{original}$.]{\includegraphics[scale = 0.28]{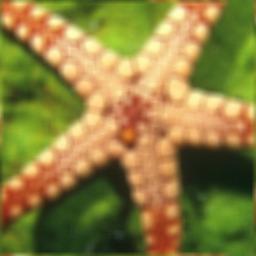}
    \includegraphics[scale = 0.28]{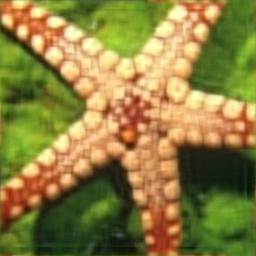} \includegraphics[scale = 0.28]{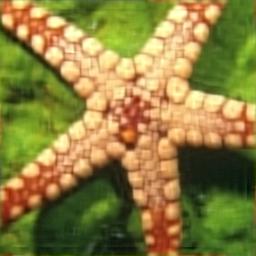} \includegraphics[scale = 0.28]{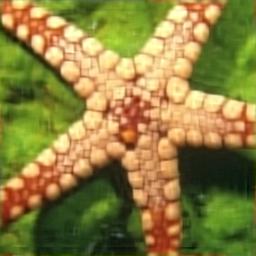} \includegraphics[scale = 0.28]{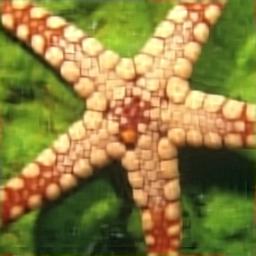} \includegraphics[scale = 0.28]{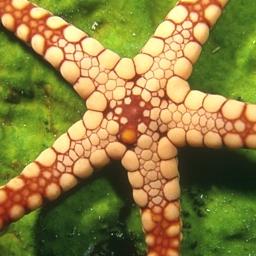}
   }
   
   \subfigure[SD-MPE: $22.56\text{dB (input)}\rightarrow 27.46\text{dB} \rightarrow  28.62\text{dB} \rightarrow 29.29\text{dB}\rightarrow 29.69\text{dB} \rightarrow \text{original}$.]{\includegraphics[scale = 0.28]{fig/resultsImjpg/input_starfish.jpg}  \includegraphics[scale = 0.28]{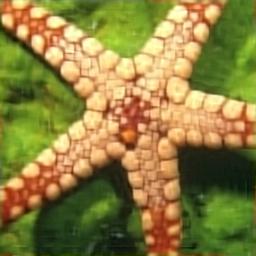} \includegraphics[scale = 0.28]{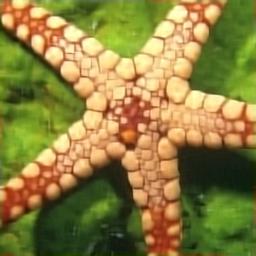} \includegraphics[scale = 0.28]{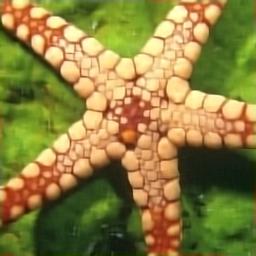} \includegraphics[scale = 0.28]{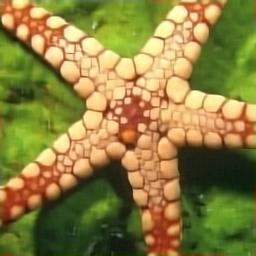} \includegraphics[scale = 0.28]{fig/resultsImjpg/starfish.jpg}}
   
   \subfigure[Nesterov: $22.56\text{dB (input)}\rightarrow 27.44\text{dB} \rightarrow  28.78\text{dB} \rightarrow 29.43\text{dB}\rightarrow 29.79\text{dB} \rightarrow \text{original}$.]{\includegraphics[scale = 0.28]{fig/resultsImjpg/input_starfish.jpg}  \includegraphics[scale = 0.28]{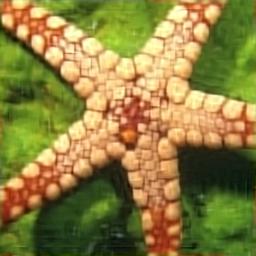} \includegraphics[scale = 0.28]{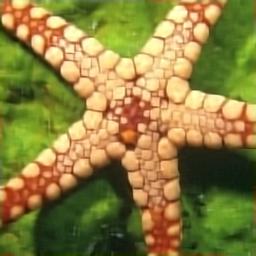} \includegraphics[scale = 0.28]{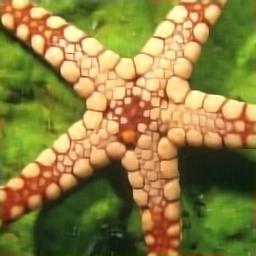} \includegraphics[scale = 0.28]{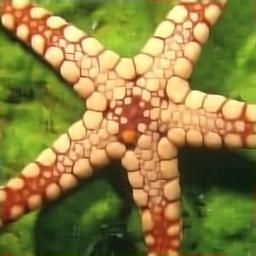} \includegraphics[scale = 0.28]{fig/resultsImjpg/starfish.jpg}}
   
     \subfigure[LBFGS: $22.56\text{dB (input)}\rightarrow 27.80\text{dB} \rightarrow  29.52\text{dB} \rightarrow 30.14\text{dB}\rightarrow 30.40\text{dB} \rightarrow \text{original}$.]{\includegraphics[scale = 0.28]{fig/resultsImjpg/input_starfish.jpg}  \includegraphics[scale = 0.28]{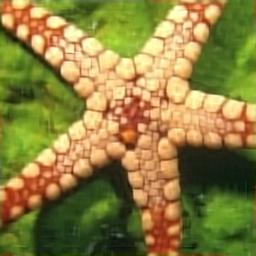} \includegraphics[scale = 0.28]{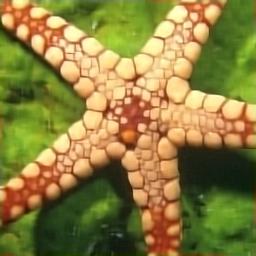} \includegraphics[scale = 0.28]{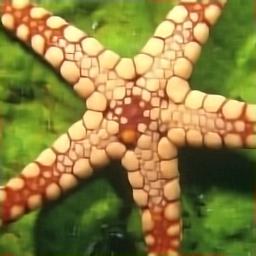} \includegraphics[scale = 0.28]{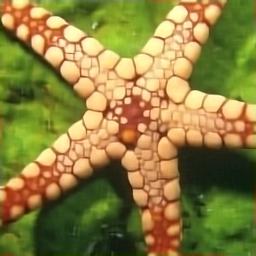} \includegraphics[scale = 0.28]{fig/resultsImjpg/starfish.jpg}}
   
    \subfigure[FP: $22.56\text{dB (input)}\rightarrow 28.91\text{dB} \rightarrow  29.76\text{dB} \rightarrow 30.13\text{dB}\rightarrow 30.31\text{dB} \rightarrow \text{original}$.]{\includegraphics[scale = 0.28]{fig/resultsImjpg/input_starfish.jpg}  \includegraphics[scale = 0.28]{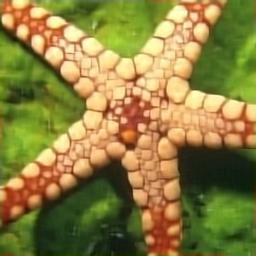} \includegraphics[scale = 0.28]{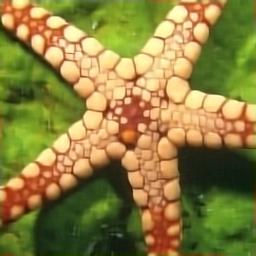} \includegraphics[scale = 0.28]{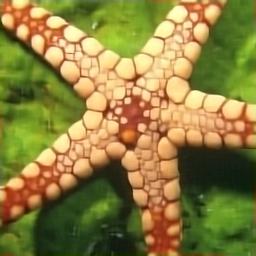} \includegraphics[scale = 0.28]{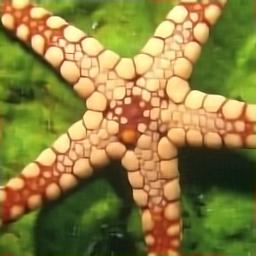} \includegraphics[scale = 0.28]{fig/resultsImjpg/starfish.jpg}}
   
    \subfigure[FP-MPE: $22.56\text{dB (input)}\rightarrow 30.07\text{dB} \rightarrow  30.55\text{dB} \rightarrow 30.60\text{dB}\rightarrow 30.60\text{dB} \rightarrow \text{original}$.]{\includegraphics[scale = 0.28]{fig/resultsImjpg/input_starfish.jpg}  \includegraphics[scale = 0.28]{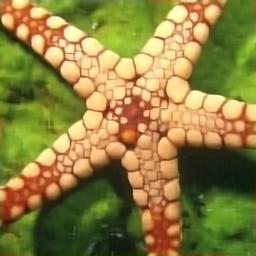} \includegraphics[scale = 0.28]{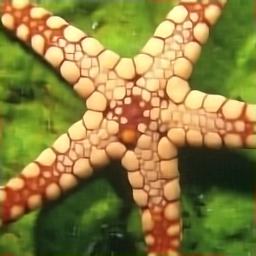} \includegraphics[scale = 0.28]{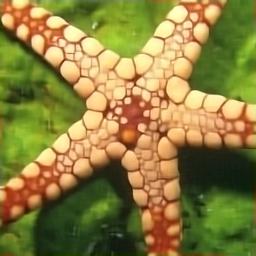} \includegraphics[scale = 0.28]{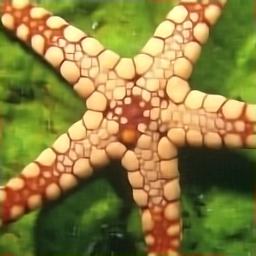} \includegraphics[scale = 0.28]{fig/resultsImjpg/starfish.jpg}}
   
    \caption{Reconstruction results of different algorithms in various iterations for the uniform kernel. From left to right: Blurred one $\rightarrow$ $\# 20$ $\rightarrow$ $\#40$ $\rightarrow$ $\#60$ $\rightarrow$ $\#80$ $\rightarrow$ Ground truth.}
    \label{fig:imageRe_qualityUniform}
\end{figure*}

\begin{figure}[!htb]
		\centering
		\subfigure[Cost value versus iteration.]{\includegraphics[scale = 0.31]{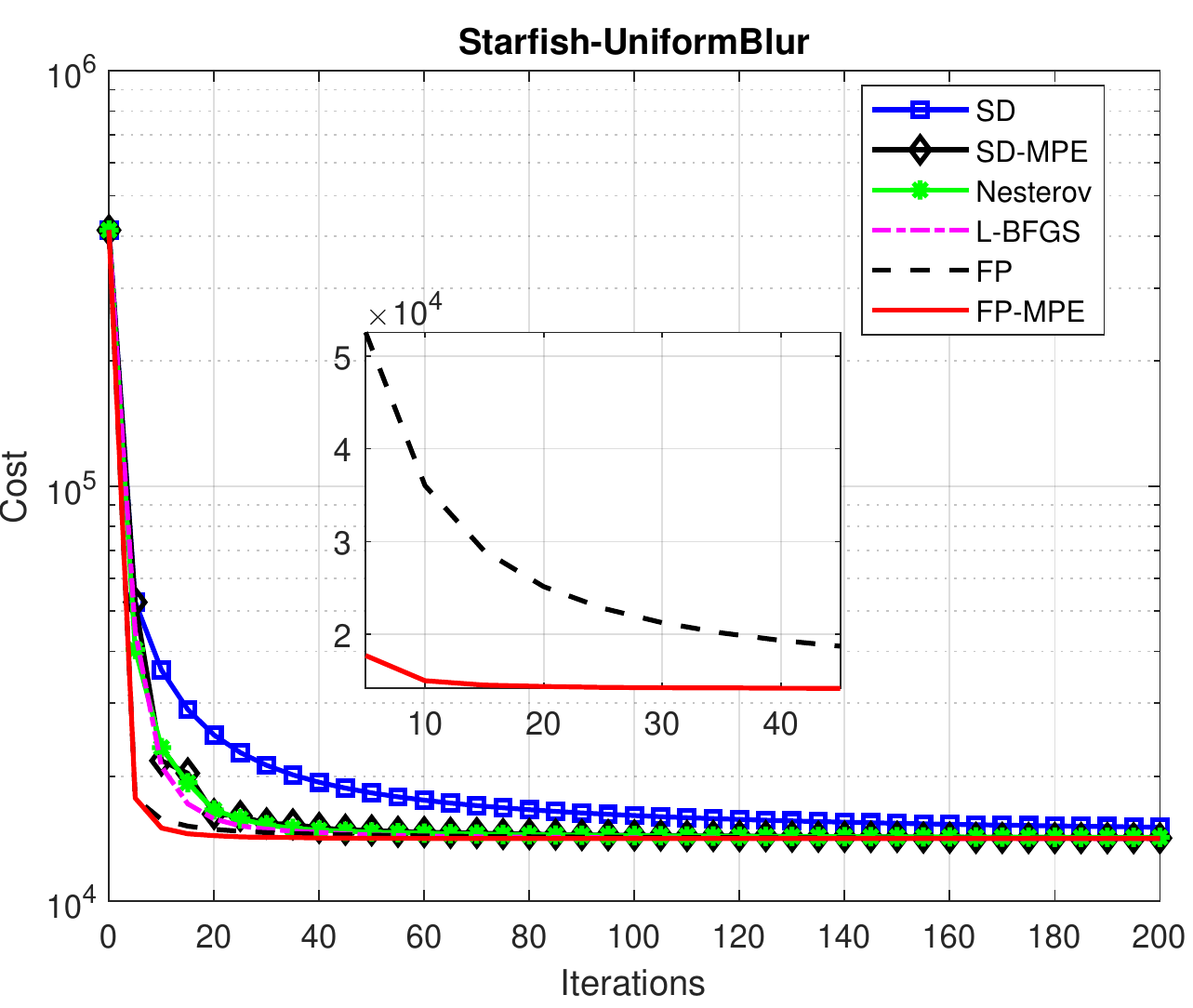}}
		\subfigure[Cost value versus CPU time.]{\includegraphics[scale = 0.31]{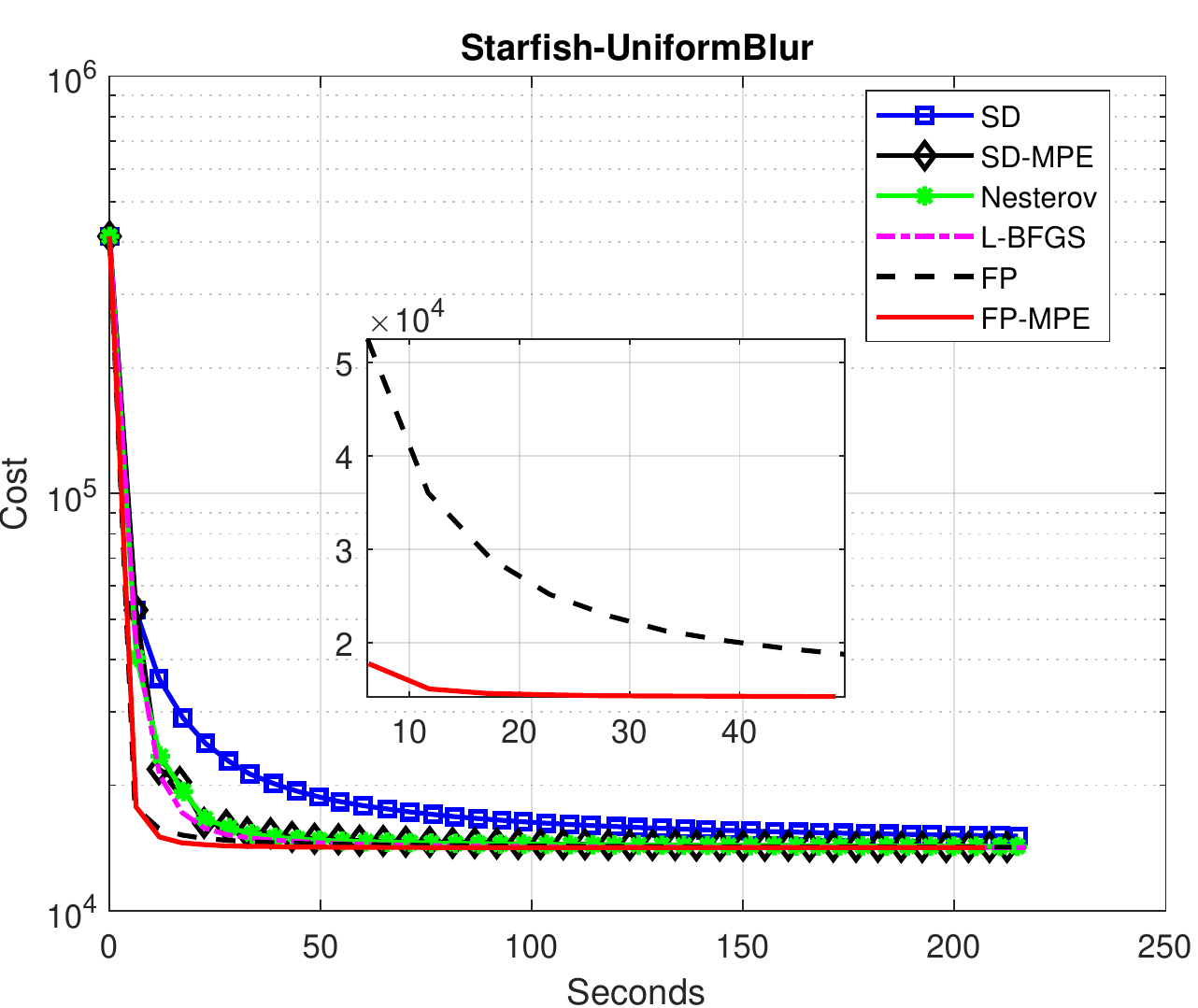}}
		
		\subfigure[PSNR versus iteration.]{\includegraphics[scale = 0.31]{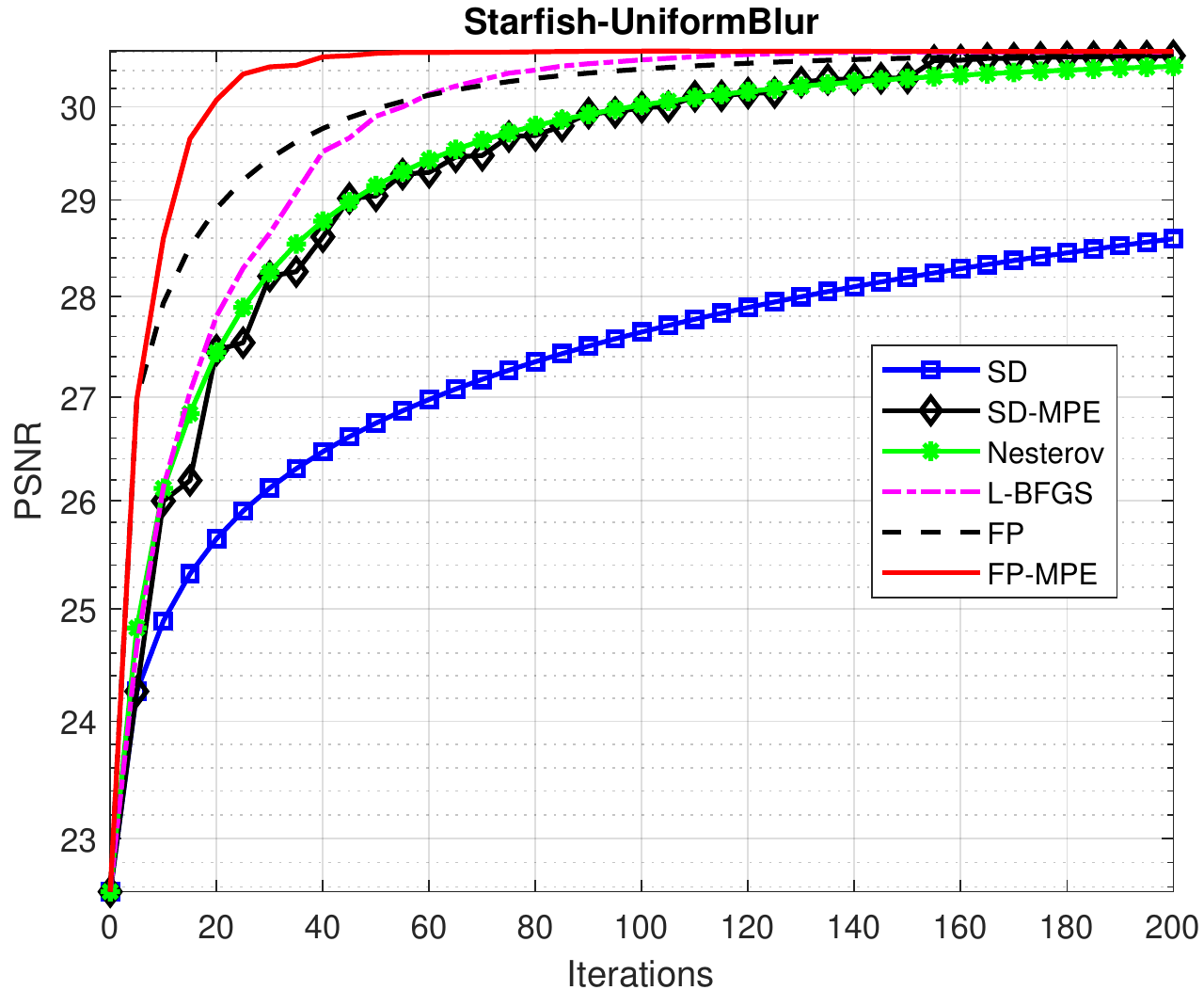}}
		\subfigure[PSNR versus CPU time.]{\includegraphics[scale = 0.31]{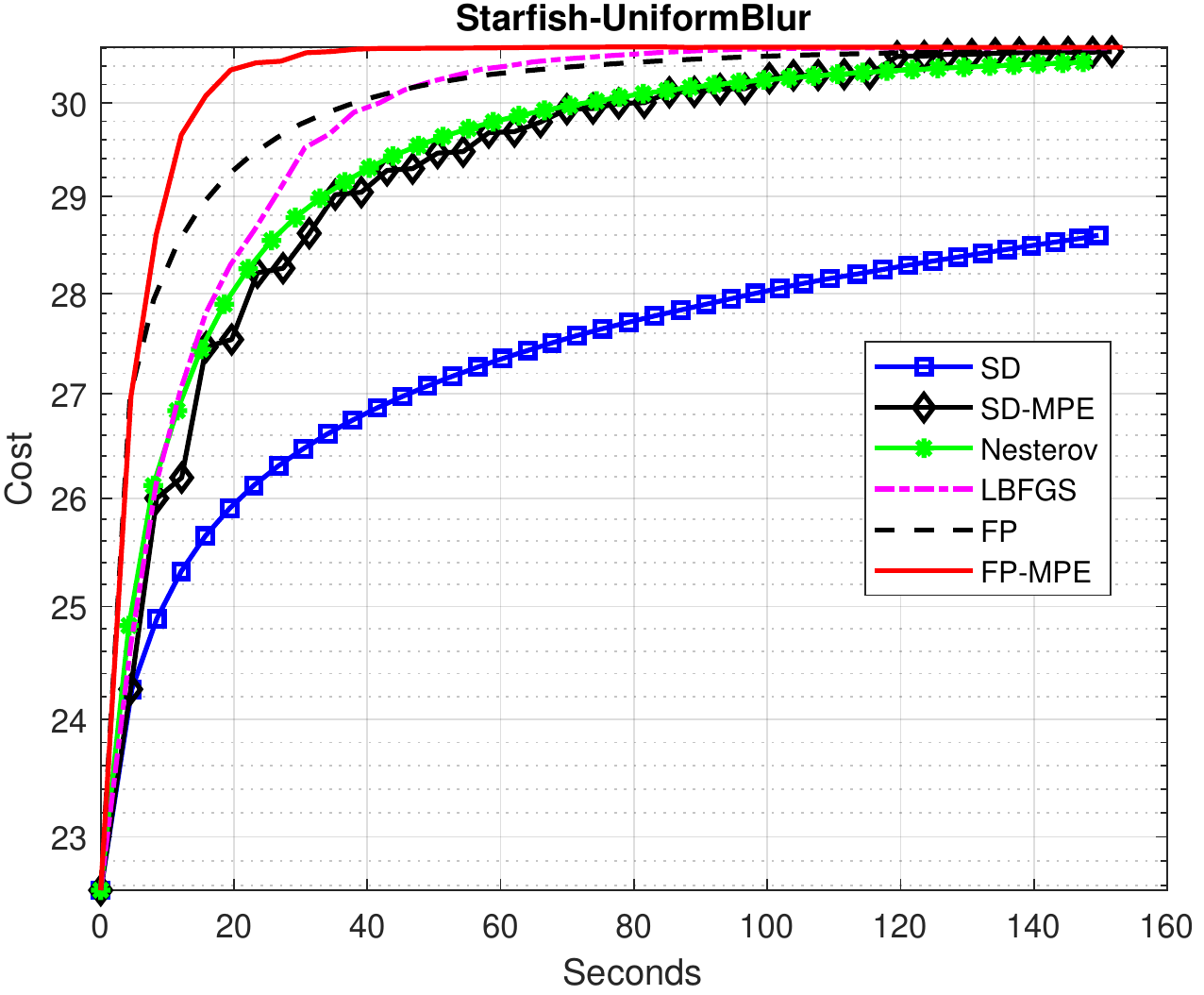}}
		\caption{Image Deblurring - Uniform Kernel, ``Starfish'' Image.}\label{Fig:deblur:starfish:uniform}
	\end{figure}

\begin{figure}[!htb]
		\centering
		\subfigure[Cost value versus iteration.]{\includegraphics[scale = 0.31]{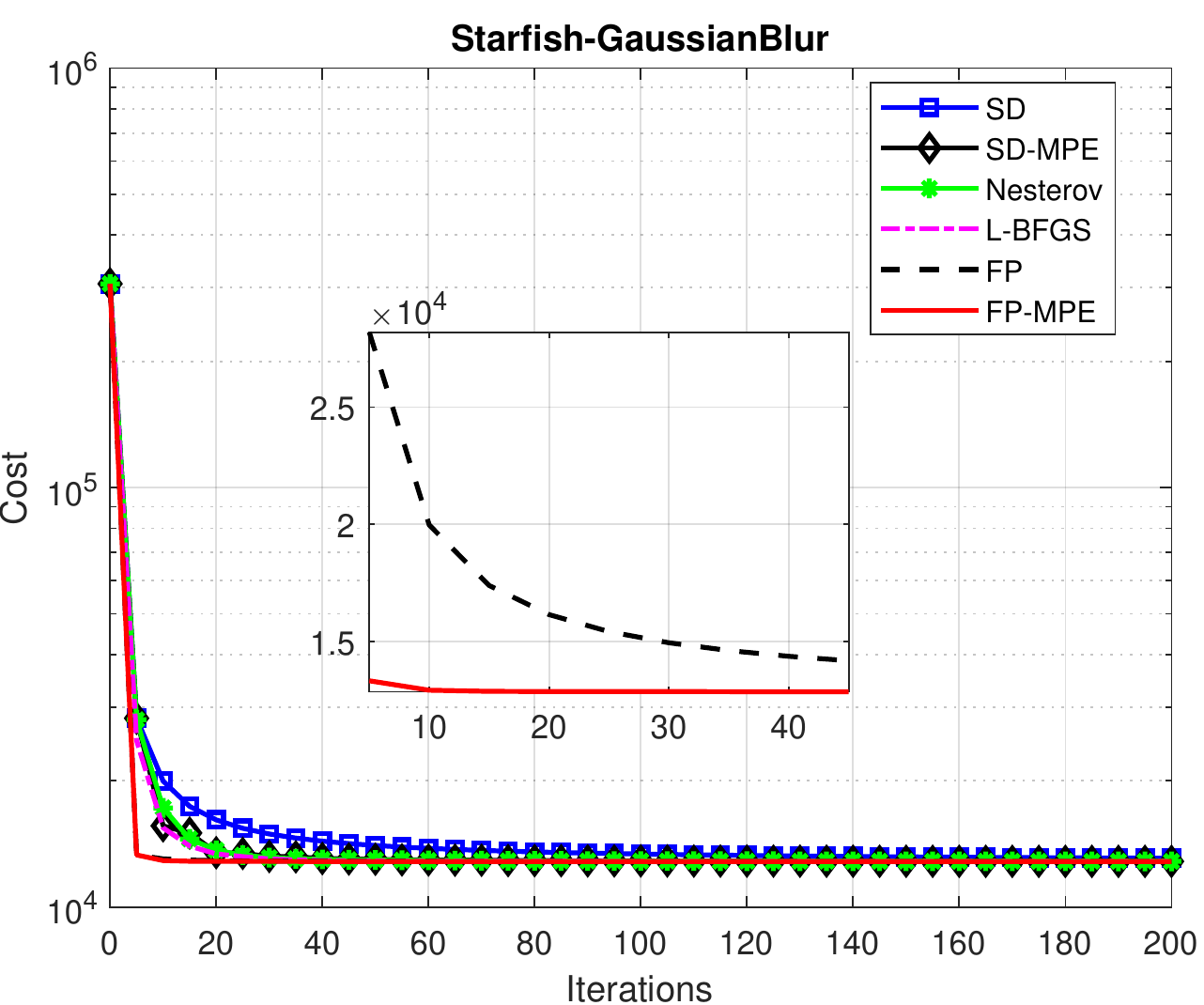}}
		\subfigure[Cost value versus CPU time.]{\includegraphics[scale = 0.31]{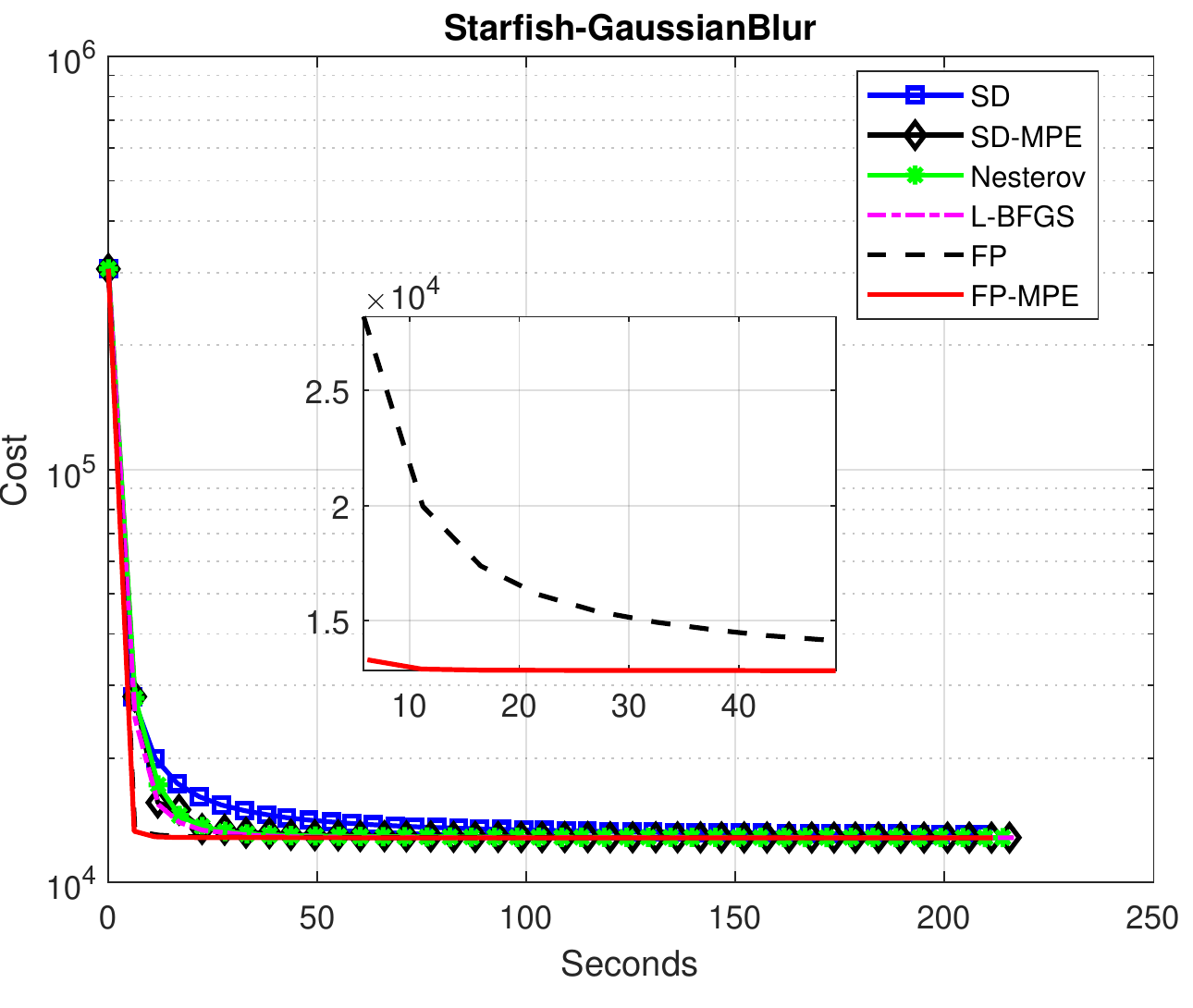}}
				
		\subfigure[PSNR versus iteration]{\includegraphics[scale = 0.31]{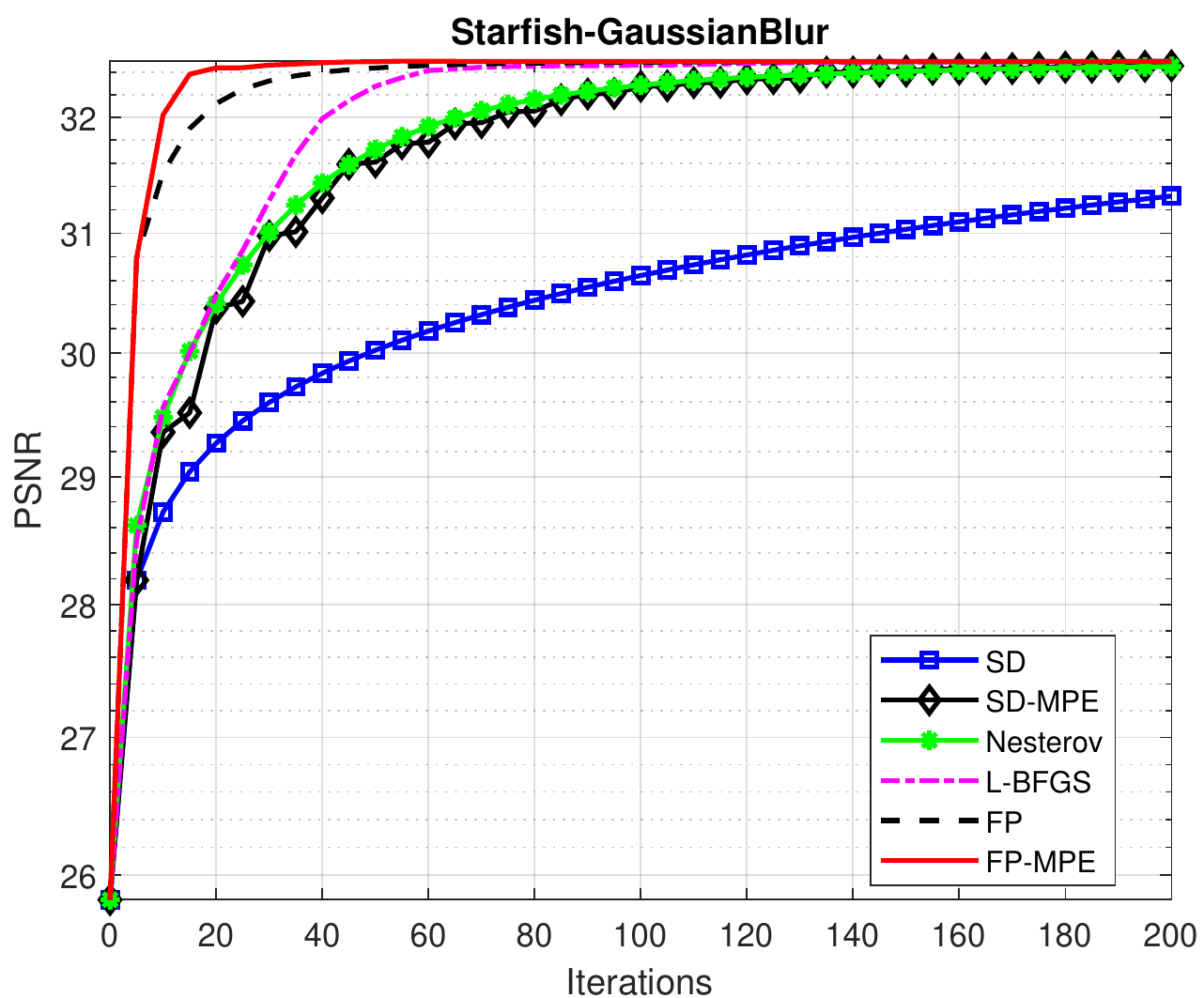}}
		\subfigure[PSNR versus iteration]{\includegraphics[scale = 0.31]{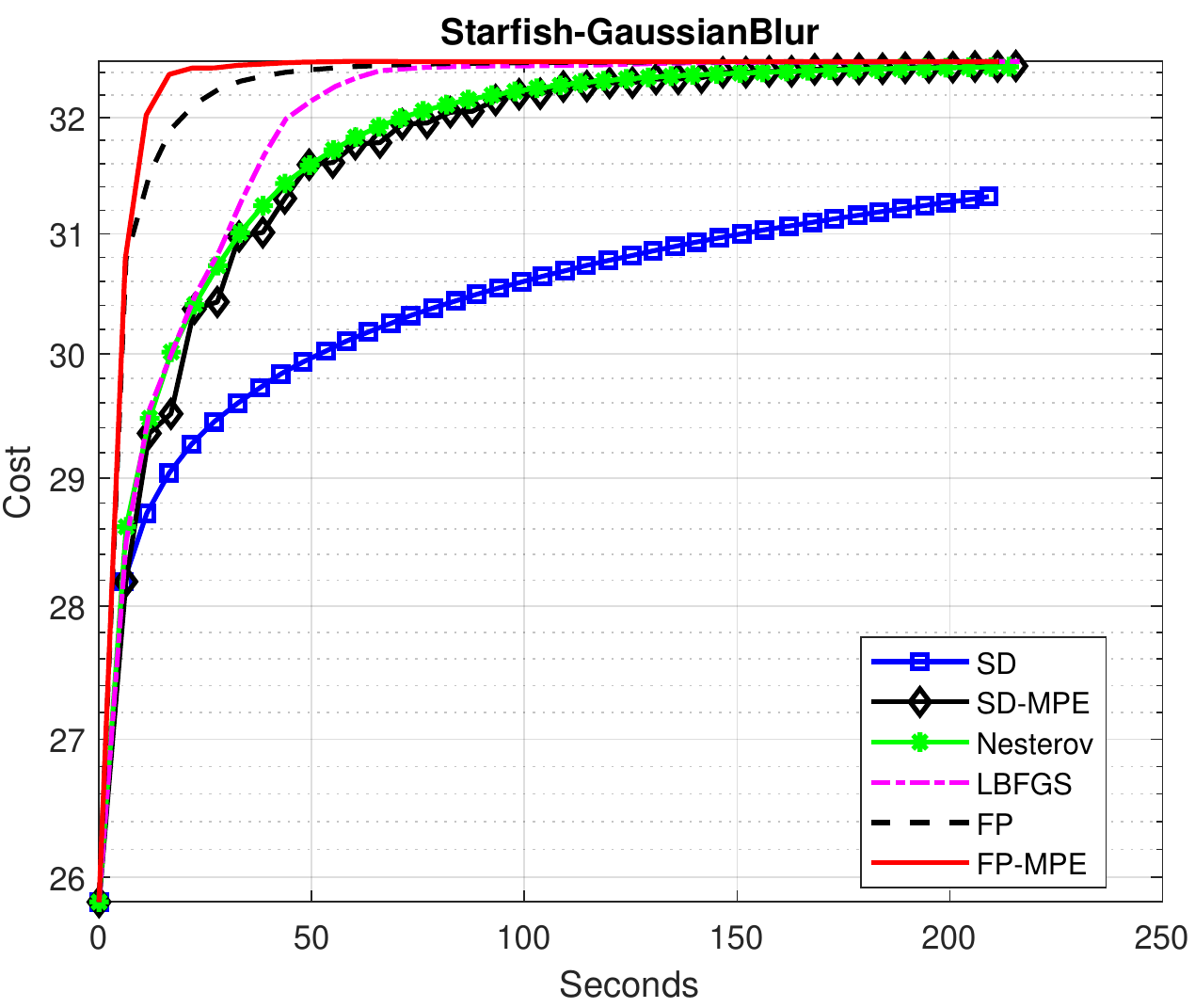}}
		\caption{Image Deblurring - Gaussian Kernel, ``Starfish'' Image. }\label{Fig:deblur:starfish:gaussian}
	\end{figure}

{\color{black} In this experiment, we degrade the test images by convolving with two different point spread functions (PSFs), i.e., $9\times 9$ uniform blur and a Gaussian blur with a standard derivation of $1.6$. In both of these cases, we add an additive Gaussian noise with $\sigma=\sqrt{2}$ to the blurred images.}
The parameters $m$ and  $\kappa$ in {\bf Algorithm \ref{alg:VE:Prac}} are set to $0$ and $5$ for the image deblurring task. {\color{black} Additionally, we apply VE to the case where the baseline algorithm is SD, called SD-MPE, with  the parameters $m$ and $\kappa$ are chosen as $0$ and $8$.} {\color{black}The value of the cost function and peak signal to noise ratio (PSNR) versus iteration or CPU time are given in Fig.s \ref{Fig:deblur:starfish:uniform} and \ref{Fig:deblur:starfish:gaussian}\footnote{{\color{black} The goal of this paper is to investigate the performance of solving RED with VE rather than the restoration results. Therefore, we present the recovered PSNR versus iteration or running time. One can utilize some no-reference quality metrics like NFERM \cite{gu2015using} and ARISMc \cite{gu2015no} to further examine the restoration results.}}}. These correspond to both a uniform and a Gaussian blur kernels, all tested on the ``starfish'' image. {\color{black} Clearly, we observe that SD is the slowest algorithm. Surprisingly, SD-MPE and Nesterov's method yield almost the same convergence speed, despite their totally different scheme. Moreover, We note that FP is faster than L-BFGS, Nesterov's method and SD-MPE.} Undoubtedly, FP-MPE is the fastest one, both in terms of iterations and CPU time, which indicates the effectiveness of MPE's acceleration.  {\color{black} To provide a visual effect, we show the change in reconstructed quality of different algorithms in Fig. \ref{fig:imageRe_qualityUniform}. Clearly, the third column of FP-MPE achieves the best reconstruction faster, while other methods need more iterations to obtain a comparable result.}

Additional nine test images suggested in \cite{romano2017little} are also included in our experiments, in order to investigate the performance of VE further. {\color{black}In this experiment we focus on the comparison between FP-MPE and FP for the additional images.}  We run the native FP method $200$ iterations first and {\color{black} denote the final image by $\vx^*$. Clearly, the corresponding cost-value is $E(\vx^*)$. We activate {\bf Algorithm \ref{alg:VE:Prac}} with the same initial value as used in the FP method to examine how many iterations are needed to attain the same or lower objective value than $E(\vx^*)$.} The final number of iterations with different images are given in Table \ref{tab:deblur:cost}.  Clearly, an acceleration is observed in all the test images in the image deblurring task.

\begin{table*}[!htb]
		\centering
		\caption{The number of iterations with different images for FP and FP-MPE in image deblurring task to attain the same cost.}
		\begin{tabular}{|c||c|c|c|c|c|c|c|c|c|c|}
			\hline
			\textbf{Image} & \textsf{Butterfly} & \textsf{Boats} & \textsf{C. Man} & \textsf{House} & \textsf{Parrot} & \textsf{Lena}  & \textsf{Barbara} & \textsf{Starfish} & \textsf{Peppers} & \textsf{Leaves}  \\
			\hline
			\multicolumn{11}{|c|}{\textbf{Deblurring: Uniform kernel, $ \sigma=\sqrt{2} $}} \\
			\hline
			
			RED: FP-TNRD & 200	& 200 & 200	& 200	& 200	& 200	& 200	& 200	& 200	& 200 \\
			
			RED: FP-MPE-TNRD & 60 & 55 & 55 & 80 & 50 & 50 & 55 & 50  & 55  & 55 \\
			\hline
			\multicolumn{11}{|c|}{\textbf{Deblurring: Gaussian kernel, $ \sigma=\sqrt{2} $}} \\
			\hline
			RED: FP-TNRD & 200	& 200	& 200	& 200	& 200	& 200	& 200	& 200	& 200	& 200	 \\
			RED: FP-MPE-TNRD & 70 & 65 & 55 & 65  & 55 & 80 & 45 & 55 & 95 & 80 \\
			\hline
			
		\end{tabular}
		
		\label{tab:deblur:cost}
	\end{table*}
	
\subsection{Image super-resolution}\label{SubSec:ImageSR}
\begin{figure}[!htb]
		\centering
		\subfigure[Cost value versus iteration.]{\includegraphics[scale = 0.31]{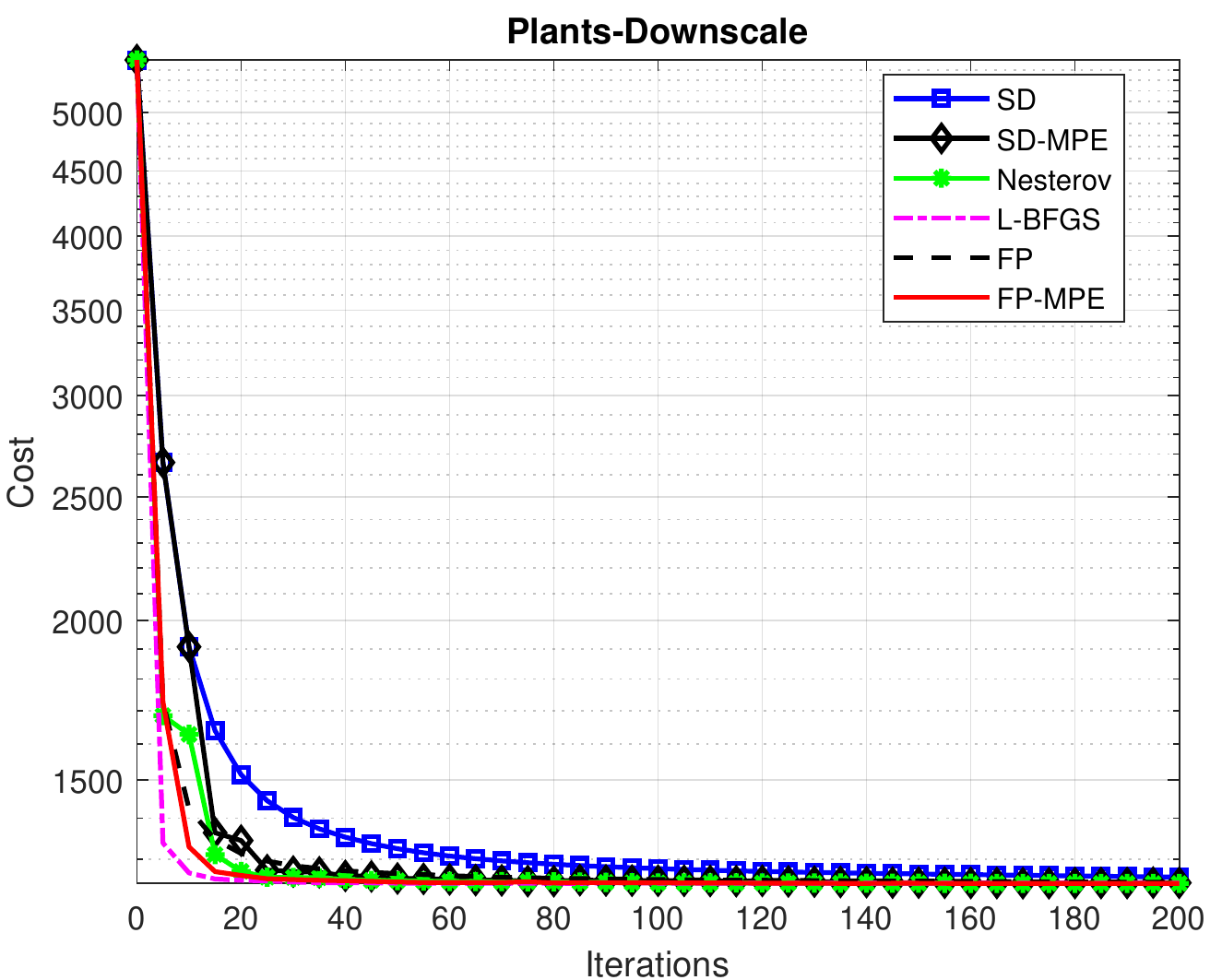}}
		\subfigure[Cost value versus CPU time.]{\includegraphics[scale = 0.31]{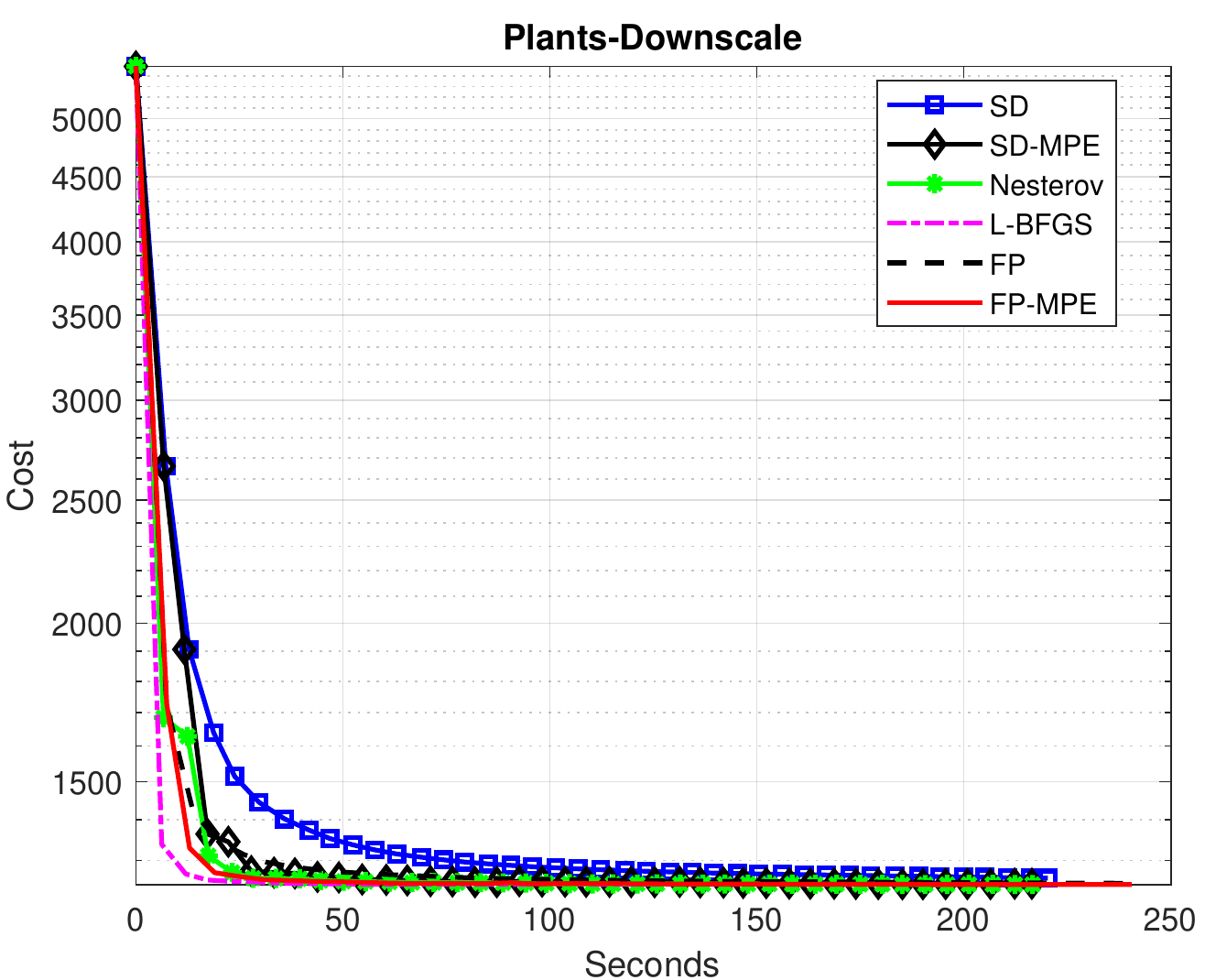}}
				
		\subfigure[PSNR versus iteration.]{\includegraphics[scale = 0.31]{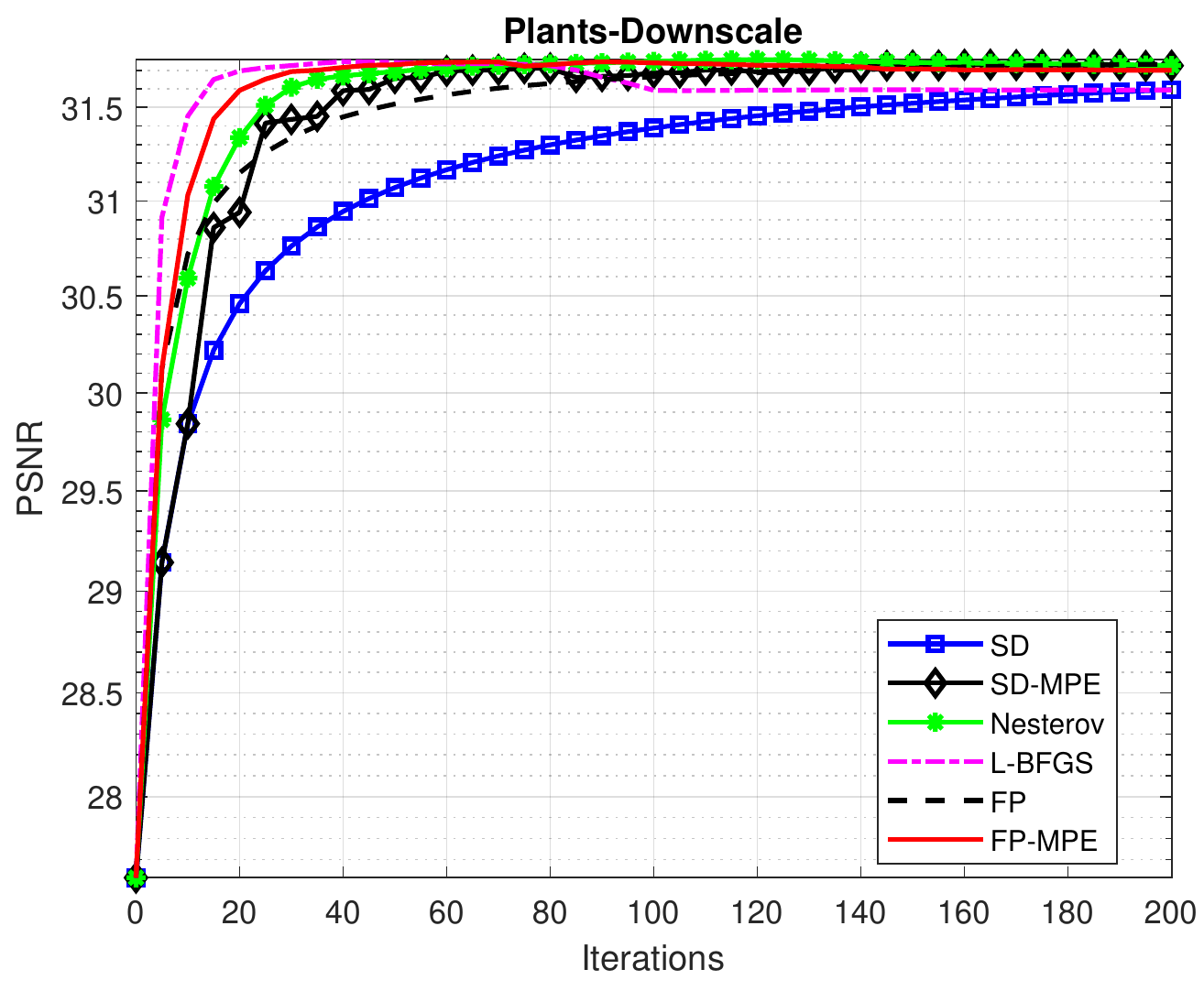}}
		\subfigure[PSNR versus CPU time.]{\includegraphics[scale = 0.31]{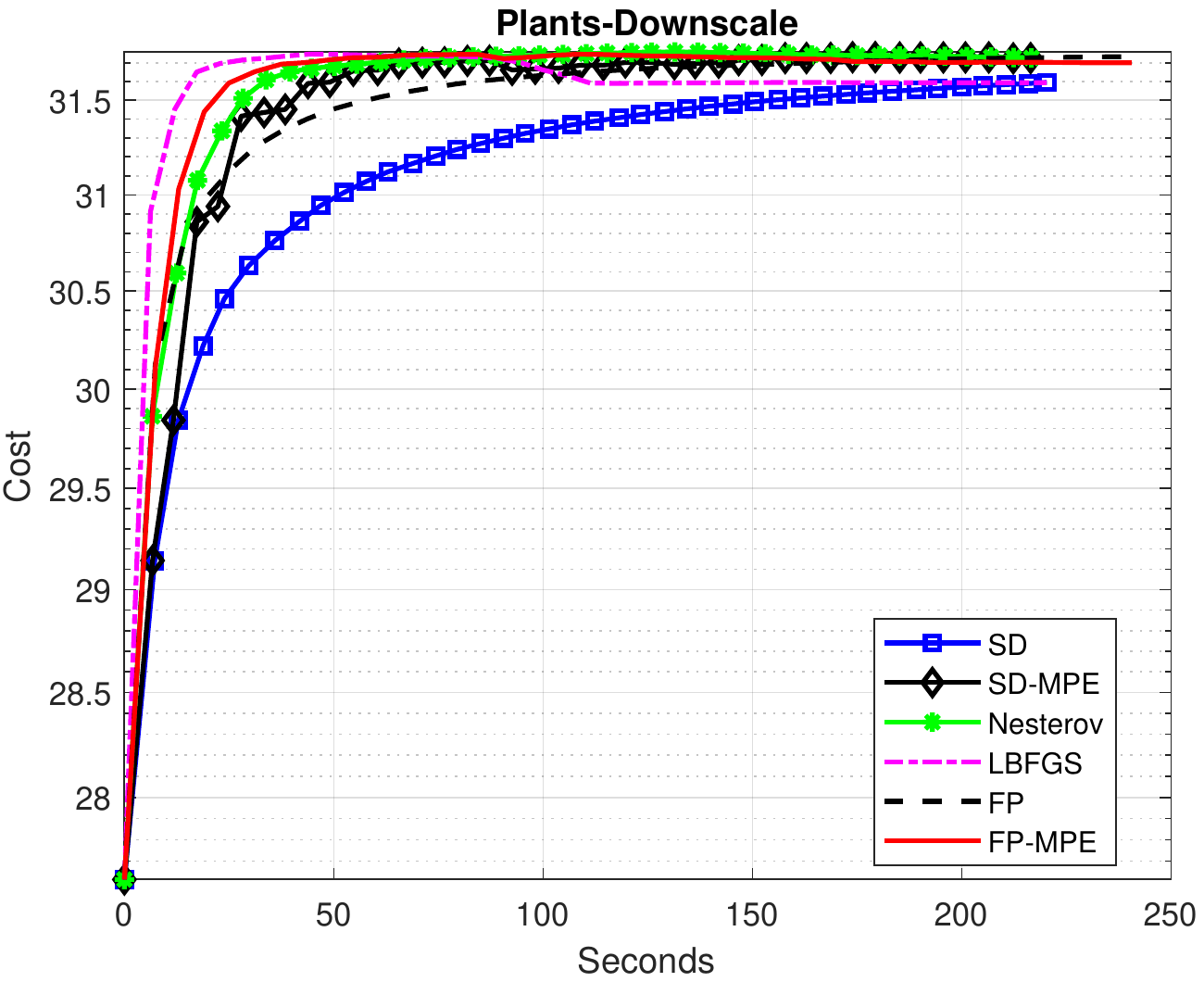}}
		\caption{Image Super-Resolution, ``Plant'' Image.}\label{Fig:SR:plants}
	\end{figure}

{\color{black} We generate a low resolution image by blurring the ground truth one with a $7\times7$ Gaussian kernel with standard derivation $1.6$ and then downsample by a factor of $3$. Afterwards, an additive Gaussian noise with $\sigma=5$ is added to the resulting image. The same parameters $m$ and $\kappa$ used in the deblurring task for FP-MPE are adopted here. For SD-MPE, the parameters $m$ and $\kappa$ are set to $1$ and $10$, respectively.} We choose ``Plants'' as our test image because it needs more iterations for FP-MPE to converge. {\color{black}As observed from Fig. \ref{Fig:SR:plants}, while L-BFGS and the Nesterov’s method are faster than the FP method, our acceleration method (FP-MPE) is quite competitive with both.} Furthermore, we investigate all of the test images as shown in \cite{romano2017little} to see how many iterations are needed for MPE to achieve the same or lower cost compared with the FP method. The results are shown in Table \ref{tab:SR:cost}. As can be seen, MPE works better than the FP method indicating an effective acceleration for solving RED. 
	
	
\begin{table*}[!htb]
	\centering
	\caption{The number of iterations with different images for FP and FP-MPE in image super-resolution task to attain the same cost.}
	\begin{tabular}{|c||c|c|c|c|c|c|c|c|c|}
			\hline
			\multicolumn{10}{|c|}{\textbf{Super-Resolution, scaling = $3$, $ \sigma= 5 $}} \\
			\hline
			\textbf{Image} & \textsf{Butterfly} & \textsf{Flower} & \textsf{Girl} & \textsf{Parth.} & \textsf{Parrot} & \textsf{Raccoon}  & \textsf{Bike} & \textsf{Hat} & \textsf{Plants} \\
			\hline
			
			RED: FP-TNRD & 200	& 200 & 200	& 200	& 200	& 200	& 200	& 200	& 200	 \\
			
			RED: FP-MPE-TNRD & 60   &  65 & 50 & 70 & 55 & 60 & 60 &50 & 70 \\
			\hline
			
		\end{tabular}
		
		\label{tab:SR:cost}
	\end{table*}
	
\subsection{The Choice of the Parameters and the Difference Among RRE, MPE and SVD-MPE} \label{SubSec:ParaRobust}
\begin{figure}[!htb]
		\centering
	\subfigure[$m=0$ and $\kappa=3,5,10$.]{\includegraphics[scale = 0.31]{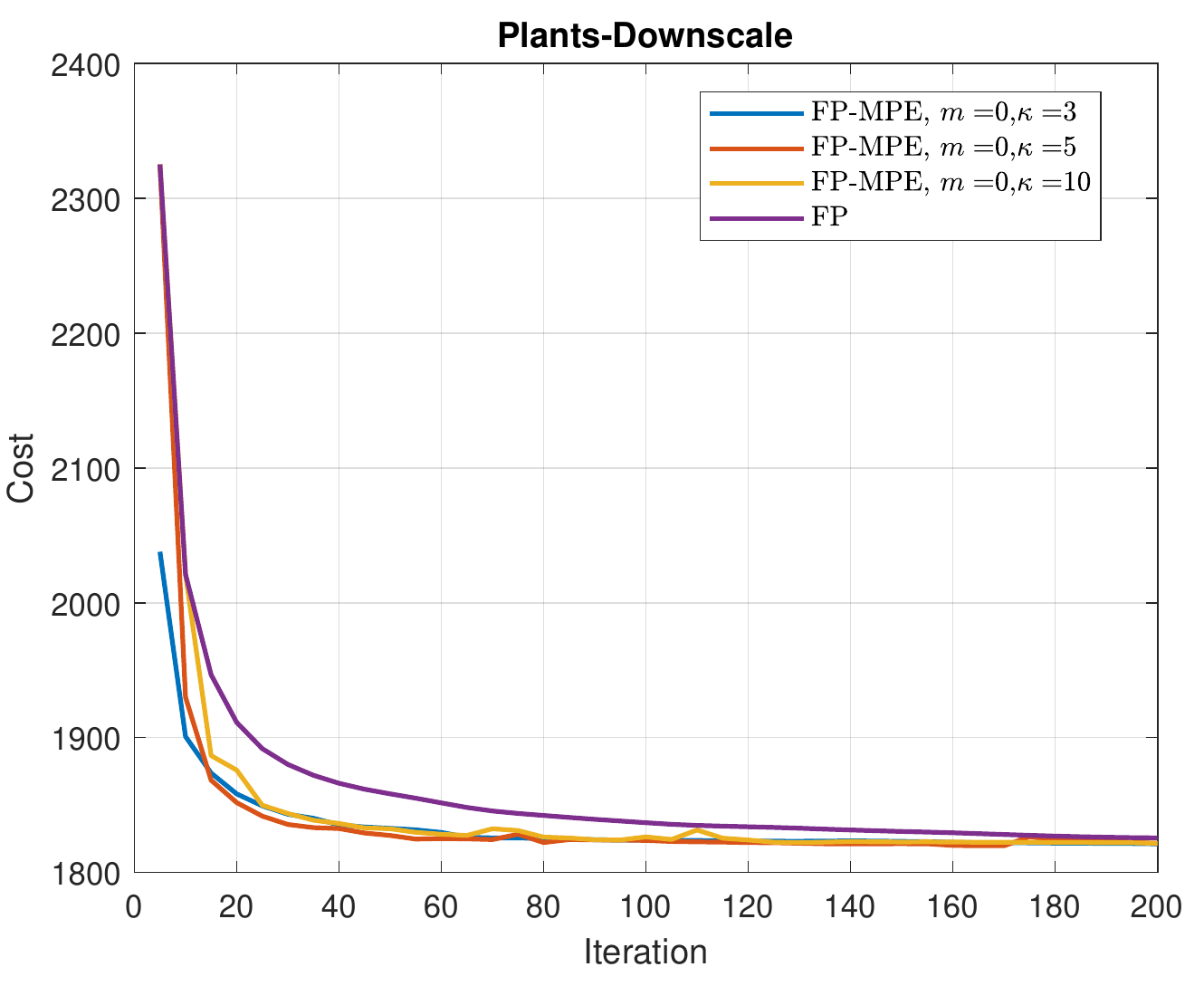}\label{Fig:SR:para:1}}
	\subfigure[$m=5$ and $\kappa=3,5,10$.]{\includegraphics[scale = 0.31]{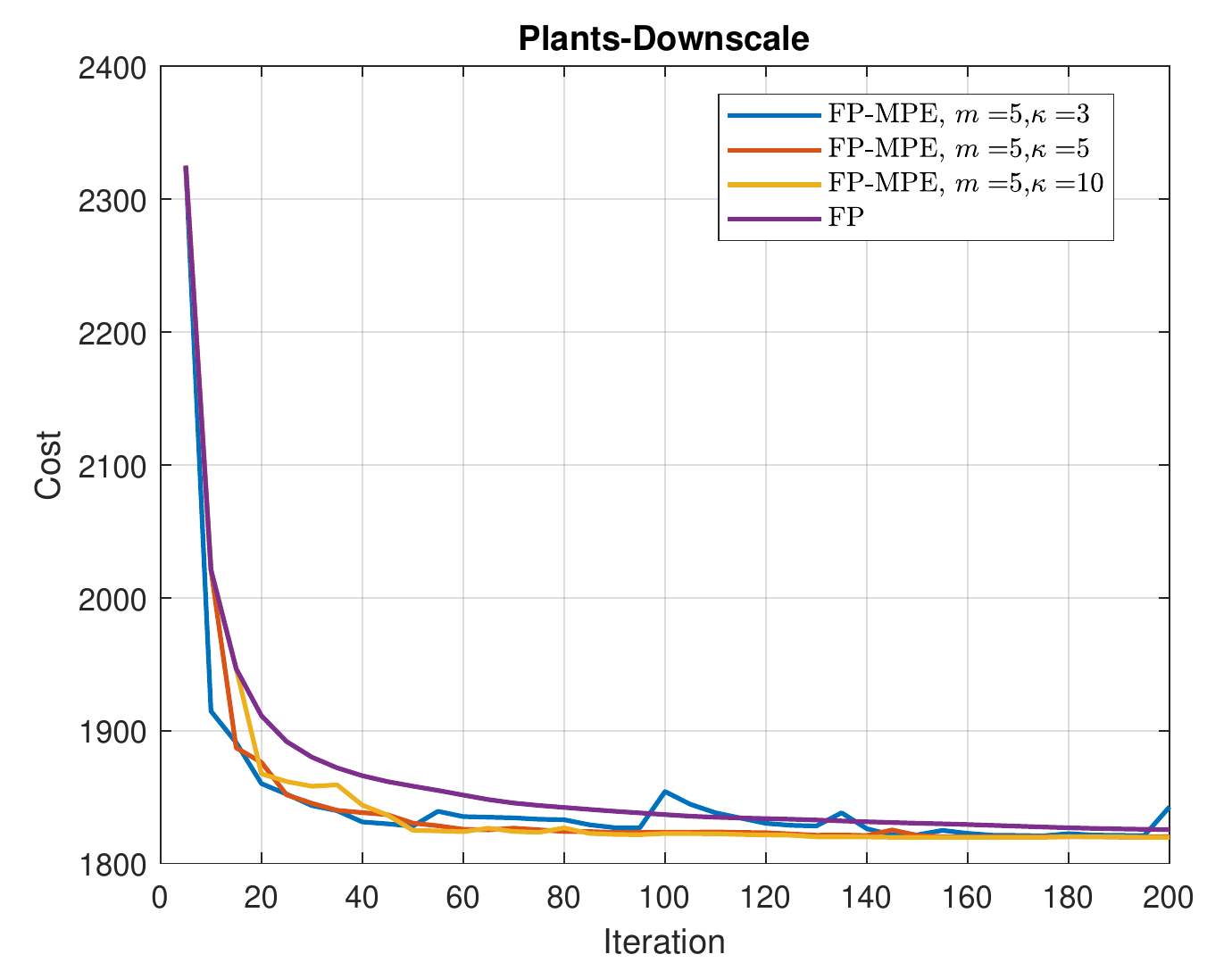}\label{Fig:SR:para:2}}
		
	\subfigure[$m=10$ and $\kappa=3,5,10$.]{\includegraphics[scale = 0.31]{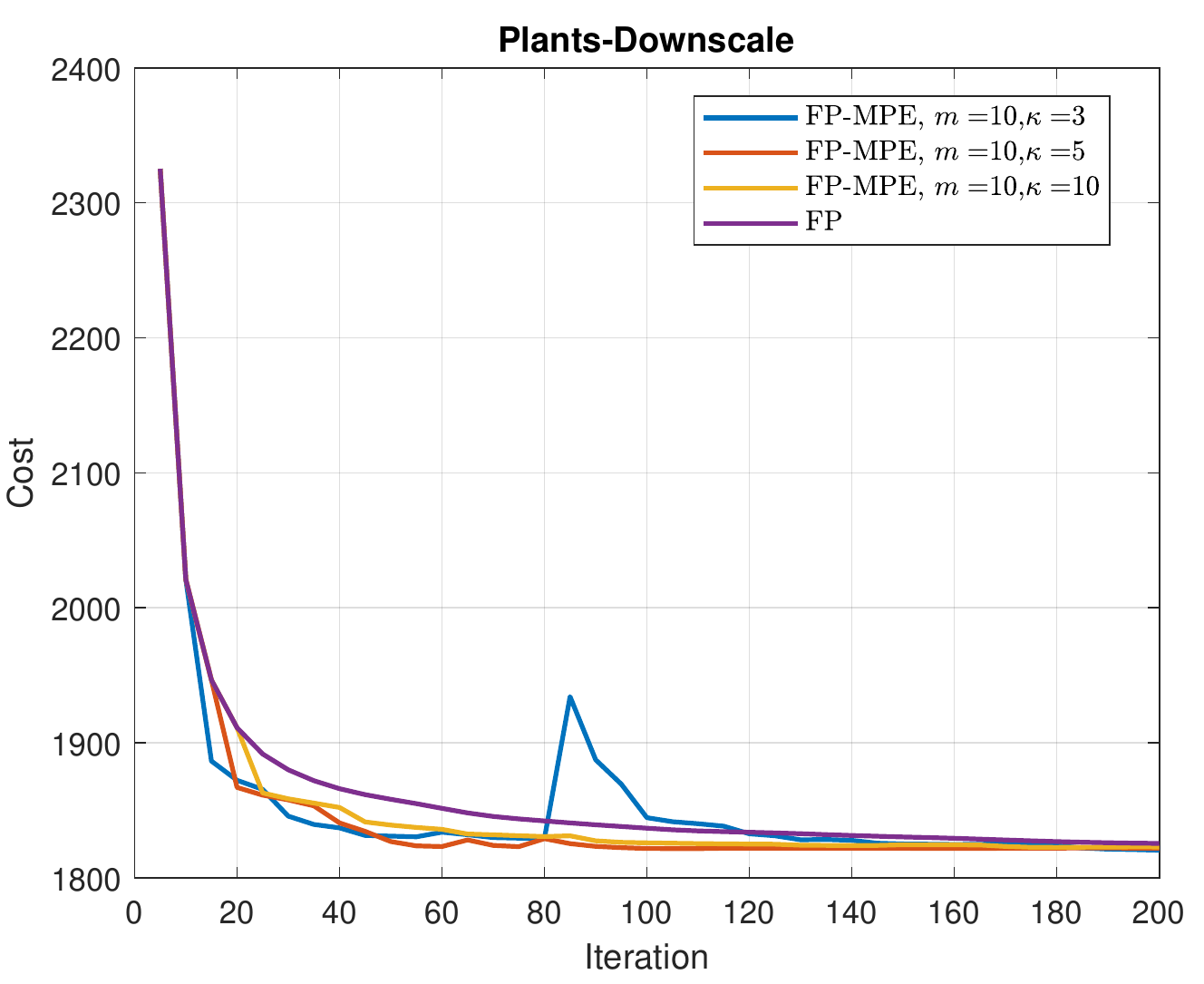}\label{Fig:SR:para:3}}
	\subfigure[Comparsion among MPE, RRE and SVD-MPE.]{\includegraphics[scale = 0.31]{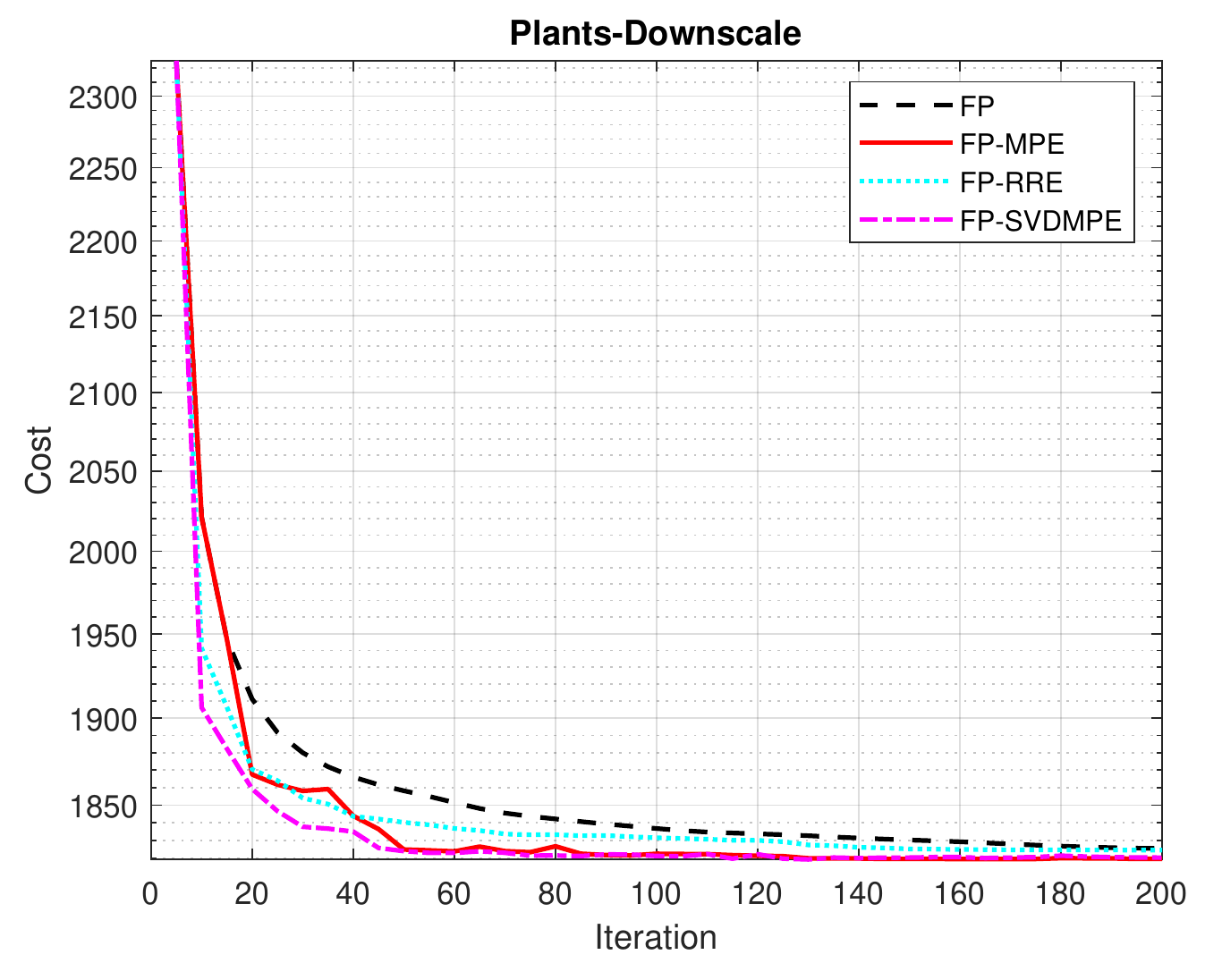}\label{Fig:SR:para:comparsion:1}}
\caption{Exploring the robustness to the choice of the parameters in MPE, and the difference among the three VE schemes. All these graphs correspond to the test image ``Plants'' for the single image super-resolution task.}\label{Fig:SR:para:comparsion}
\end{figure}

Conclude by discussing the robustness to the choice of parameters $m$ and $\kappa$ for the MPE algorithm. To this end, the single image super-resolution task is chosen as our study. Furthermore, we choose to demonstrate this robustness on the ``Plants'' image since it required the largest number of iterations in the MPE recovery process. As seen from Fig.s \ref{Fig:SR:para:1} - \ref{Fig:SR:para:3}, MPE always converges faster than the regular FP method with different $m$ and $\kappa$. Moreover, we also observe that a lower objective value is attained through MPE. Notice that MPE has some oscillations because it is not a monotonically accelerated technique. However, we still see a lower cost is achieved if additional iterations are given. 

In part (d) of Fig. \ref{Fig:SR:para:comparsion}, an optimal pair of $m$ and $\kappa$ is chosen for MPE, RRE and SVD-MPE for the single image super-resolution task with the ``Plants'' image.\footnote{The optimal $m$ and $\kappa$ are obtained by searching in the range $[0,10]$ with $\kappa\geq2$, seeking the fastest convergence for these three methods.} We see that all three methods yield an acceleration and a lower cost, demonstrating the effectiveness of the various variants of VE. Moreover, we see that SVD-MPE converges faster at the beginning, but MPE yields a lowest eventual cost. 
    

\section{Conclusion}\label{Sec:Conclusion}
The work reported in \cite{romano2017little} introduced RED -- a flexible framework for using arbitrary image denoising algorithms as priors for general inverse problems. This scheme amounts to iterative algorithms in which the denoiser is called repeatedly. While appealing and quite practical, there is one major weakness to the RED scheme -- the complexity of denoising algorithms is typically high which implies that the use of RED is likely to be costly in run-time. This work aims at deploying RED efficiently, alleviating the above described shortcoming. An accelerated technique is proposed in this paper, based on the Vector Extrapolation (VE) methodology. The proposed algorithms are demonstrated to substantially reduce the number of overall iterations required for the overall recovery process. We also observe that the choice of the parameters in the VE scheme is robust.

	\bibliographystyle{ieeetr}
	\bibliography{References_AccelRed}
	
	%
	%
	%

\end{document}